\documentclass[preprint,12pt]{elsarticle}

\usepackage{amsmath}

\usepackage{amsfonts}

\usepackage[ruled, vlined]{algorithm2e}
\usepackage{algorithmic}
\usepackage{array}
\usepackage{textcomp}
\usepackage{stfloats}
\usepackage{url}
\usepackage{verbatim}
\usepackage{graphicx}
\usepackage{subfigure}

\usepackage[hidelinks,colorlinks,citecolor=green]{hyperref}

\usepackage{cleveref}
\usepackage{makecell}
\usepackage{threeparttable}
\usepackage{diagbox}

\usepackage{booktabs}

\usepackage{pifont}
\usepackage{paralist}
\usepackage{relsize}

\usepackage{xcolor}
\usepackage{graphicx}
\usepackage{amssymb}
\usepackage{caption}

\usepackage{framed,multirow}

\usepackage{times}
\usepackage{soul}

\usepackage{amsthm}

\usepackage{enumerate}

\usepackage[switch]{lineno}

\def\sw{\texttt{RL-I2IT}}

\journal{Neural Networks}

\begin{document}

\begin{frontmatter}



\title{RL-I2IT: Image-to-Image Translation with  Deep Reinforcement Learning}



\author[1]{Jing Hu} 
\author[2]{Ziwei Luo}
\author[1]{Chengming Feng}
\author[3]{Shu Hu}
\author[4]{Bin Zhu}
\author[1]{Xi Wu}
\author[5]{Xin Li}
\author[6]{Hongtu Zhu}
\author[7]{Siwei Lyu}
\author[8]{Xin Wang \corref{cor1}} \ead{xwang56@albany.edu}

\cortext[cor1]{Corresponding author.}

\affiliation[1]{organization={School of Computer Science},
            addressline={Chengdu University of Information Technology}, 
            city={Chengdu},
            postcode={610225}, 
            state={Sichuan},
            country={China}}           
\affiliation[2]{organization={Department of Information Technology},
            addressline={Uppsala University}, 
            city={Uppsala},
            postcode={75105}, 
            state={Uppsala},
            country={Sweden}}
\affiliation[3]{organization={School of Applied and Creative Computing},
            addressline={Purdue University}, 
            city={West Lafayette},
            postcode={46202}, 
            state={IN},
            country={USA}} 
\affiliation[4]{organization={Microsoft Research Asia},
            addressline={No. 5 Danling Street, Tower 1, First Floor, Haidian District}, 
            city={Beijing},
            postcode={100080}, 
            state={Beijing},
            country={China}} 
\affiliation[5]{organization={Department of Computer Science},
            addressline={University at Albany (SUNY)}, 
            city={NY},
            postcode={12222}, 
            state={NY},
            country={USA}}
\affiliation[6]{organization={Departments of
Biostatistics, Statistics, Computer Science, and Genetics},
            addressline={University of North Carolina at Chapel Hill}, 
            city={Chapel Hill},
            postcode={27514}, 
            state={North Carolina},
            country={USA}}
\affiliation[7]{organization={Department of Computer Science and Engineering},
            addressline={University at Buffalo (SUNY)}, 
            city={NY},
            postcode={12222}, 
            state={NY},
            country={USA}}
\affiliation[8]{organization={College of Integrated Health Sciences, and AI Plus Institut},
            addressline={University at Albany (SUNY)}, 
            city={NY},
            postcode={12222}, 
            state={NY},
            country={USA}}

\begin{abstract}
Most existing Image-to-Image Translation (I2IT) methods generate images in a single run of deep learning (DL) models. 
However, designing a single-step model often requires many parameters and suffers from overfitting. Inspired by the analogy between diffusion models and reinforcement learning, we reformulate I2IT as an iterative decision-making problem via deep reinforcement learning (DRL) and propose a computationally efficient RL-based I2IT (\sw) framework. 
The key feature in the \sw~framework is to decompose a monolithic learning process into small steps with a lightweight model to progressively transform the source image to the target image. 
Considering the challenge of handling high-dimensional continuous state and action spaces in the conventional RL framework, we introduce meta policy with a new ``concept Plan'' to the standard Actor-Critic model. This plan is of a lower dimension than the original image, which facilitates the actor to generate a tractable high-dimensional action. 
In the \sw~framework, we also employ a task-specific auxiliary learning strategy to stabilize the training process and improve the performance of the corresponding task.
Experiments on several I2IT tasks demonstrate the effectiveness and robustness of the proposed method when facing high-dimensional continuous action space problems. Our implementation of the \sw~framework is available at 
\url{https://github.com/lesley222/RL-I2IT}.
\end{abstract}


\end{frontmatter}



\section{Introduction}
\label{sec:introduction}
Many computer vision problems, such as face inpainting, semantic segmentation, 
realistic photo generated from sketch, and neural style transfer, can be unified under the framework of learning image-to-image translation (I2IT) \cite{pang2021image}.
Existing approaches to I2IT can be categories into either one-step deep-learning (DL) framework  (e.g., Variational Autoencoders \cite{kingma2013auto}, U-Net \cite{ronneberger2015u}, and conditional GANs \cite{isola2017image}) or iterative diffusion models (e.g., Palette \cite{saharia2022palette}, SSDM \cite{sun2023sddm}, Plug-and-Play \cite{tumanyan2023plug}). 
Directly learning I2IT with one-step DL models typically suffers from two major challenges. 
One is that to handle high-dimensional I2IT problems, one-step DL models typically have complex structures and many parameters, making them difficult to train and hard to deploy in resource-limited scenarios such as mobile devices.
The other is that many of these models do not generalize well~\cite{neyshabur2017exploring} due to the abundance of global minima caused by the over-parameterized setting. 
Although these problems can be potentially alleviated by using multi-scale models or multi-stage pipelines such as diffusion models, we still face the challenge of prohibitive computational complexity. 

\begin{figure}[t]
  \centering
  \includegraphics[width=1\linewidth]{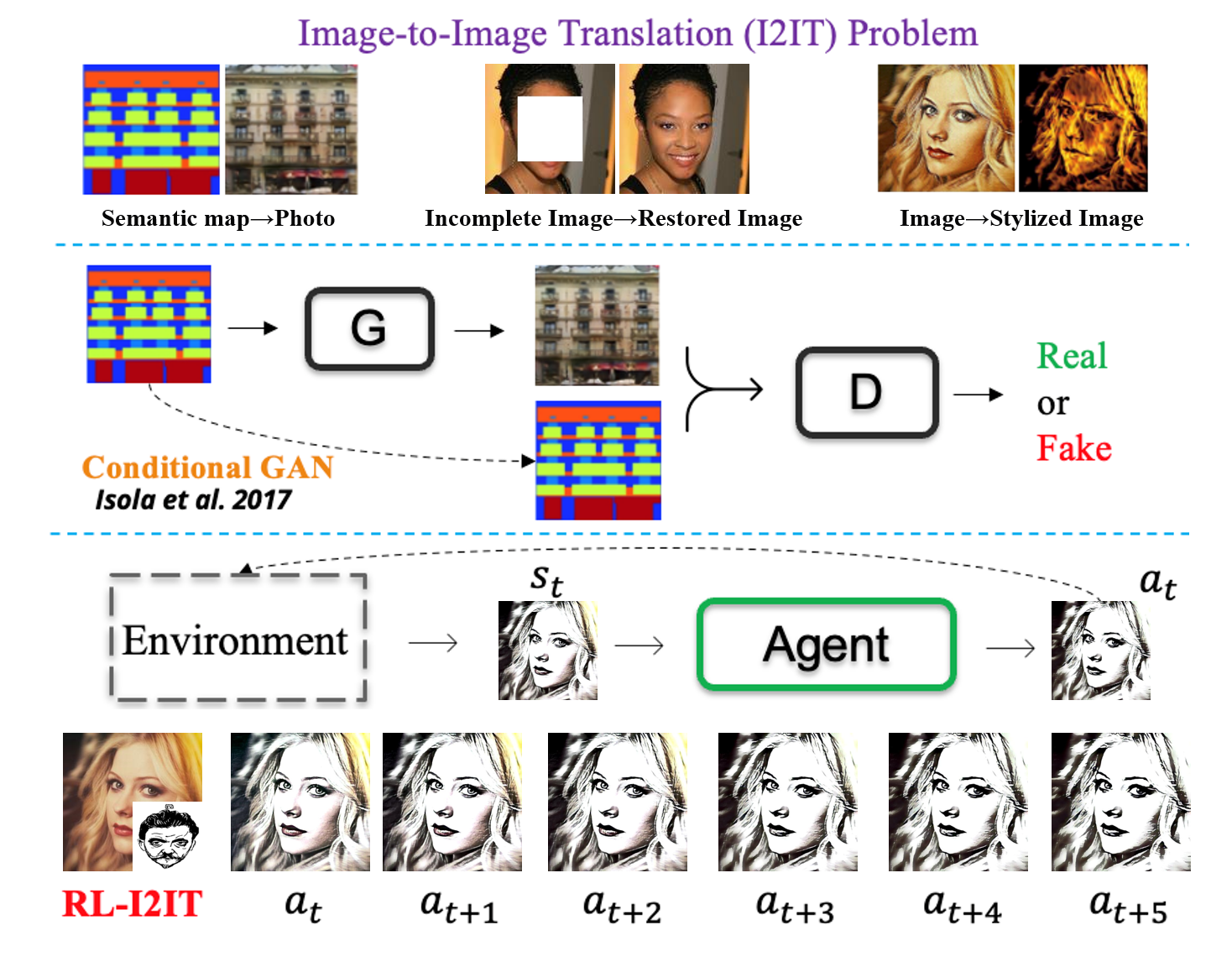}
  \caption{\small \textbf{Top:} I2I Problem translates an image from a source domain to a target domain. \textbf{Mid:} Example of one-step method CGAN \cite{isola2017image}. \textbf{Bottom:} Our RL-based stepwise I2IT progressively transforms the source image, and the process is demonstrated clearly.}
  \label{fig_introduction}
\end{figure}

To address these limitations 
, we explore solving I2IT problems by leveraging the recent advances in deep reinforcement learning (DRL). 
The key idea is to decompose the monolithic learning process into small steps with a lightweight CNN, aiming to improve the quality of predicted results progressively (See Fig. \ref{fig_introduction}). 
By decomposing a one-step complex task into a series of simpler tasks, our approach 
can handle the simplified task with a much simpler network rather than using a large, heavily parameterized network. 
%
Although recent works have successfully applied DRL to solve several visual tasks~\cite{caicedo2015active,hu2021end,luo2020spatiotemporal}, their action spaces are usually discrete, making them unsuitable for I2IT, which requires continuous action spaces. 
A promising direction for learning continuous actions is the maximum entropy reinforcement learning (MERL), which improves 
exploration
by maximizing a standard RL objective with an entropy term~\cite{haarnoja2018soft}. Soft actor-critic (SAC)~\cite{haarnoja2018soft} is an instance of MERL and has been applied to solve continuous action tasks \cite{xiang2022rmbench}. 
However, the main issue hindering the applicability of SAC on I2IT is its inability to handle high-dimensional states and actions effectively. 
Recently, RAE~\cite{yarats2019improving} tried to address this problem by combining SAC with a regularized autoencoder, but it only provides an auxiliary loss for end-to-end RL training and is incapable of the I2IT tasks. 
Besides, high-dimensional
states and actions require training an I2IT-based RL model to make much more exploration and exploitation, which leads to unstable training~\cite{yarats2019improving}.
One solution to stabilize training is to extract a lower-dimensional visual representation with a separately pre-trained DNN model and learn the
value function and corresponding
policy in the latent space~\cite{nair2018visual}. 
However, this approach can not be trained from scratch. Otherwise, it can lead to inconsistent state representations with an optimal policy. 



\begin{figure*}[t]
    \centering
    \includegraphics[scale=0.39]{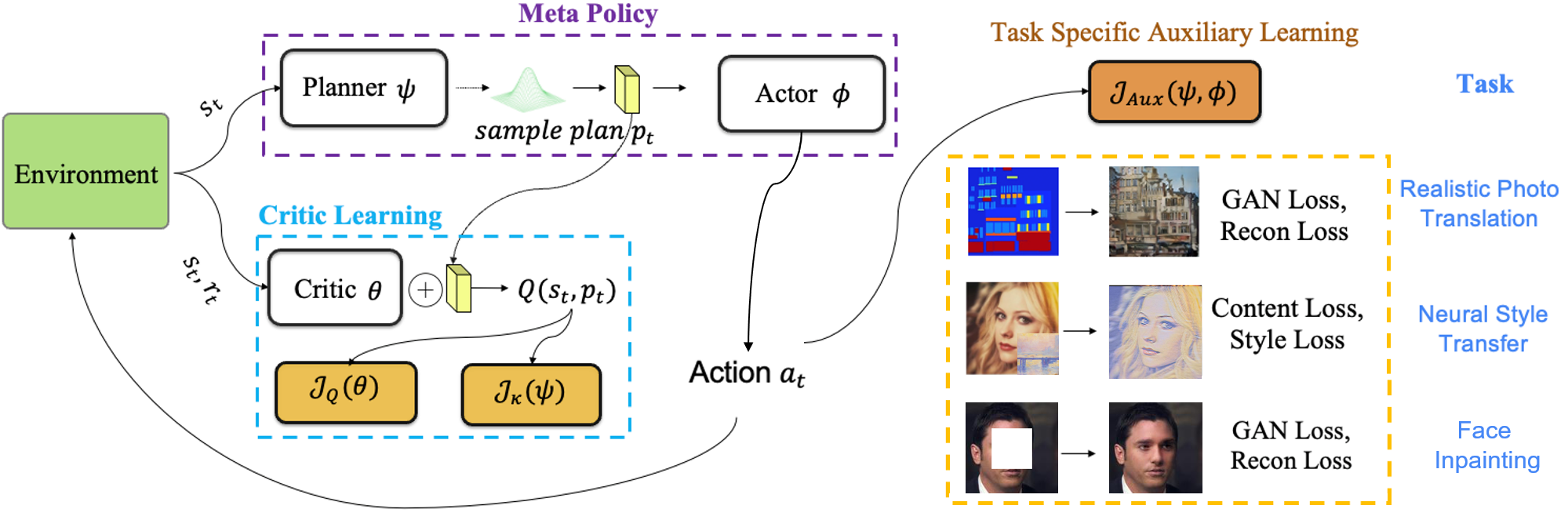}
    \vspace{-0.2cm}
    \caption{
    Our \sw~framework with a Planner-Actor-Critic structure. \textbf{Left:} At time step $t$, the  environment receives executable action ${\bf a}_t$, and outputs state and reward (${\bf s}_t, r_t$). In our meta policy, latent plan ${\bf p}_t$ is sampled from the planner to guide the actor to generate executable action ${\bf a}_t$ that interacts with the environment. The plan is also evaluated by the critic. The nature of \({\bf a}_t\) is also task-dependent, 
    for tasks aiming at realistic image generation, such as face inpainting or neural style transfer, \({\bf a}_t\) could directly be the target image. 
    \textbf{Right:} Task-specific auxiliary learning objectives depend on specific tasks for various purposes, such as stabilizing the training process or improving performance.
    }
    \label{ovspac}
    \vspace{-0.5cm}
\end{figure*}

Inspired by the analogy between diffusion models and RL \cite{black2023training}, we propose a new DRL framework, named \sw, for I2IT problems to handle high-dimensional continuous state and action spaces.
As shown in Fig.~\ref{ovspac}, the \sw\ framework comprises three core deep neural networks: a planner, an actor, and a critic.
We introduce a new ``concept plan'' to decompose the decision-making process into two steps, state $\rightarrow$ plan and plan $\rightarrow$ action. We call this process meta policy. The plan  
is a subspace of appropriate actions based on the current state. It is not applied to the state directly. Instead, it is used to guide the actor to generate a tractable high-dimensional action that interacts with the environment. The plan can be considered as an intermediate transition between state and action. As the input of the actor, the plan has a much lower dimension compared with the state, making it easier for the actor to learn to predict actions. Meanwhile, the plan can be evaluated by the critic efficiently since the Q function is easier to learn in the low-dimensional latent space.
Furthermore, compared with training a one-step differentiable DL-based model, it is much harder to learn from such a high-dimensional continuous control problem with traditional RL frameworks. To address it, we also employ a task-specific auxiliary learning strategy to stabilize the training process and improve the performance of the corresponding task. 
The auxiliary learning part could be any learning technology that is flexible and can readily leverage any other advanced losses or objectives.
For example, we use the standard $L_2$ reconstruction loss as auxiliary learning in many I2IT tasks. 
Our main contributions can be summarized as follows:
\begin{compactitem}

\item A new DRL framework \sw\ is proposed to handle the complex I2IT problem with high-dimensional continuous actions by decomposing the monolithic learning process into small steps.

\item To tackle the high-dimensional continuous action learning problem, we propose a stochastic meta policy that divides the decision-making processing into two steps: state $\rightarrow$ low-dimensional plan and plan $\rightarrow$ action. The plan guides the actor to predict a tractable action, and the critic evaluates the plan. The approach makes the whole learning process feasible and computationally efficient. 

\item Compared to existing DL-based models, our DRL-based model is lightweight, making it simple and computationally efficient. For example, compared to a recent one-step I2IT model pix2pixHD of size 45.9M \cite{wang2018high}, the size of our model is only 9.7M. 

\item  Our \sw~framework is flexible in incorporating many advanced auxiliary learning methods for various complex I2IT applications. Experimental results on a variety of applications, 
from face inpainting to neural style transfer
show that our approach achieves state-of-the-art performance. 


\end{compactitem}

This paper extends our previous conference papers~\cite{luostochastic} and \cite{fengijcai23} 
substantially in the following aspects: 
(i) We propose a general RL-based framework for the I2IT problem. In this regard, our previous work \cite{luostochastic} 
can be considered as a special case of the general framework in this paper. 
(ii) We provide more technical details for each application of the \sw~framework, such as the detailed network architectures.
(iii) We provide more diagnostic experiments for each application to demonstrate the effectiveness of our \sw~framework in computer vision applications. 
In the neural style transfer task, we extended style transfer to arbitrary style transfer task and demonstrated more experiments to demonstrate the effectiveness of our method at different resolutions.

The remaining content of this paper is organized as follows. After introducing the background in Section~\ref{relatedwork}, we describe the \sw~framework for step-wise I2IT in Section~\ref{method}. 
In Section \ref{app_cv} 
, we demonstrate experimentally the effectiveness and robustness of the \sw~framework on computer vision applications (
face inpainting \ref{exp_fi}, realistic image translation \ref{exp_rpt}, and neural style transfer \ref{exp_nst}). 
Due to space constraints, applications of our \sw~framework on additional tasks, including digit transform and 
more comparative experiments in style transfer,
are provided in the supplementary material.
We conclude the paper in Section~\ref{conclusion} with discussions of the limitations of the framework and future works.


\section{Background}
\label{relatedwork}

\subsection{Image-to-Image Translation}

Image-to-image translation (I2IT) aims to translate input images from a source domain to a target domain, such as generating realistic photos from semantic segmentation labels~\cite{isola2017image}, synthesizing completed visual targets from images with missing regions~\cite{pathak2016context}, neural style transfer~\cite{gatys2015neural}, 
etc. 
Autoencoder is leveraged in most research works to learn this process by minimizing the reconstruction error between the predicted image and the target. In addition, the generative adversarial network (GAN) is also vigorously studied in I2IT to synthesize realistic images~\cite{isola2017image}.
Subsequent works enhance I2IT performance by using a coarse-to-fine deep learning framework~\cite{yu2018generative} that recursively sets the output of the previous stage as the input of the next stage. In this way, the I2IT task is transformed into a multi-stage, coarse-to-fine solution. Although the recursion can be infinitely applied in practice, it is limited by the increasing model size and training instability. 
For more I2IT-related works, refer to a recent survey paper~\cite{pang2021image}.

More recently, diffusion models have found successful applications in many vision tasks including I2IT \cite{saharia2022palette,li2023bbdm,sun2023sddm}. In Palette \cite{saharia2022palette}, diffusion models (DM) outperform strong GAN and regression baselines on four I2IT tasks without task-specific hyper-parameter tuning, architecture customization, or any auxiliary loss. This work has inspired several DM-based approaches to I2IT such as Brownian Bridge Diffusion Model (BBDM) \cite{li2023bbdm} and score-decomposed diffusion models (SSDM) \cite{sun2023sddm}. Inspired by the success of vision-language models, text-driven I2IT based on plug-and-play diffusion features \cite{tumanyan2023plug} has shown high fidelity to input structure and scene layout, while significantly changing the perceived semantic meaning of objects
and their appearance.

\subsection{Reinforcement Learning with Continuous Action}

RL is described by an infinite-horizon Markov decision process (MDP), defined by the tuple $({\cal S},{\cal A}, {\cal U},{r}, {\gamma})$, where ${\cal S}$ is a set of states, ${\cal A}$ is action, ${\cal U}: {\cal S}  \times {\cal S} \times {\cal A} \rightarrow [0, \infty)$ represents the state transition probability density given state ${\bf s} \in {\cal S}$ and action ${\bf a} \in {\cal A}$, ${r}: {\cal S} \times {\cal A} \rightarrow \mathbb{R}$ is the reward emitted from each transition, and $\gamma \in [0,1]$ is the reward discount factor.
Standard RL learns to maximize the expected sum of rewards from the episodic environments under the trajectory distribution ${\rho}_{\pi}$.

Maximum Entropy RL (MERL) incorporates an entropy term with the policy, and the resulting objective is defined as $\sum \nolimits_{t = 1}^T \mathbb{E}_{({\bf s}_t, {\bf a}_t) \sim {\rho}_{\pi}} \left[r_t ({\bf s}_t, {\bf a}_t) + \alpha {\cal H}(\pi_\phi (\cdot | {\bf s}_t))\right]$, where $\alpha$ is a temperature parameter controlling the balance of the entropy ${\cal H}$ and the reward $r_t$.   
MERL model has proven stable and powerful in low dimensional continuous action tasks, such as games and robotic controls \cite{haarnoja2018soft}.
However, when facing complex visual problems such as I2IT, where observations and actions are high-dimensional, it remains a challenge for MERL models \cite{lee2020slac}.
Soft actor-critic (SAC) \cite{haarnoja2018soft} has been shown as a promising framework for learning continuous actions, which is an off-policy
actor-critic method that uses the above entropy-based framework to derive the soft policy iteration. 
The advantage of SAC is that it provides sample efficient learning and stability. It can improve both the exploration and robustness of the learned model. The original SAC paper reports its performance on continuous control tasks with up to 21 dimensions, which is far from enough for handling I2IT tasks. 
Recent studies~\cite{lee2020slac,yarats2019improving} have shown that the SAC has limitations when handling high-dimensional states and actions. 

More recently, stochastic latent actor-critic (SLAC) \cite{lee2020slac} improves the SAC by learning representation spaces with a latent variable model which is more stable and efficient for complex continuous control tasks. It can improve both the exploration and robustness of the learned model. 
However, the capability of SLAC is limited in a continuous action space. The reason is that the latent state representation in SLAC is only used to facilitate the training of the critic, which cannot handle tasks with a high-dimensional action space. 

\section{Reinforcement Learning for I2IT}
\label{method}

\subsection{Problem Formulation}



In our study, Image-to-Image Translation (I2IT) is reformulated as a multistep decision-making problem, where the transformation from an input image to a target image is not executed in a single step. Instead, we introduce a lightweight Deep Reinforcement Learning (DRL) model that incrementally performs the transformation, allowing the progressive addition of new details. We conceptualize I2IT as a Markov Decision Process (MDP), where translation, denoted as \({\cal T}\), moves from the current state \({\bf s}\) to the target \({\bf y}\) through a defined policy. This approach allows for a more delicate and progressive process of image transformation within the MDP framework, which can be formulated as follows.

\vspace{-1em}
\begin{equation*}
\centering
\label{multi-steps}
    {\cal T}({\bf s}) = {\cal T}_t \circ {\cal T}_{t-1} \circ \dotsm {\cal T}_0({\bf s})       = {\bf y},
\vspace{-0.6em}
\end{equation*}

where $\circ$ is a composition operator, ${\cal T}_t$ is the $t$-th translation step, which can predict the image from state ${\bf s}_t$. Moreover, state ${\bf s}$ can be defined according to the specific I2IT task.

\subsection{Stochastic Meta Policy of Planning and Acting}
\label{sec:metaplicy}







Our \sw~framework is designed to handle high-dimensional continuous states and actions in an infinite-horizon Markov Decision Process (MDP). It incorporates a novel component, a planner, specifically for continuous plan space. This approach diverges from traditional policies that directly map environmental states to actions \cite{sutton2018reinforcement}. Instead, it bifurcates the mapping process into two distinct steps: first state $\rightarrow$ plan, and then plan $\rightarrow$ action. We term this new two-step mapping process as a ``meta policy'', which allows for more intricate and layered decision-making compared to standard reinforcement learning models. 
The new MDP for our \sw~can be represented by the tuple $({\cal S},{\cal P},{\cal A},{\cal U},{r}, {\gamma})$. ${\cal S}$ is a set of states, ${\cal P}$ is a continuous plan, ${\cal A}$ is a continuous action, and ${\cal U}: {\cal S} \times {\cal P}  \times {\cal S} \times {\cal A}  \rightarrow [0, \infty)$ represents the state transition probability density of the next state ${\bf s}_{t+1}$ given state ${\bf s}_t \in {\cal S}$, plan ${\bf p}_t \in {\cal P}$ and action ${\bf a}_t \in {\cal A}$.

Our \sw~framework is shown in Fig.~\ref{ovspac}. It comprises three core deep neural networks: the planner, the actor, and the critic with parameters $\psi$, $\phi$, and $\theta$, respectively. 
The planner aims to generate a high-level plan in low-dimensional latent space to guide the actor. 
In some sense, the plan can be seen as action clusters or action templates, which are high-level crude actions. 
Unlike classic policy models, the input of the actor is a stochastic plan instead of the state. That is, the generated plan is forwarded to the actor further to create the high-dimensional action in our meta-policy model. Meanwhile, this plan is evaluated by the critic. 
By using the meta policy and the stochastic planner-actor-critic structure, \sw~ makes the learning process of a complex I2IT task easier.

Formally supposing a meta policy is defined as ($\kappa,\pi$). The stochastic plan is modeled as a subspace of the deformation field that gives a low-dimensional vector ${\bf p}_t$ based on the state ${\bf s}_t$, while action ${\bf a}_t$ is determined by the plan ${\bf p}_t$. 
Consider a parameterized planner $ \kappa_{\psi}$ and actor $\pi_{\phi}$, the stochastic plan is sampled as a representation: ${\bf p}_t \sim \kappa_{\psi}( {\bf p}_t|{\bf s}_t)$, and the action is generated by decoding the plan vector ${\bf p}_t$ into a high-dimensional executable action: ${\bf a}_t = \pi_{\phi}( {\bf a}_t|{\bf p}_t)$.  In practice, we reparameterize the planner and stochastic plan jointly using a neural network approximation ${\bf p}_t = f_\psi({\bf \epsilon}_t,{\bf s}_t)$, known as the reparameterization trick~\cite{kingma2013auto}, where ${\bf \epsilon}_t$ is an input noise vector sampled from a fixed Gaussian distribution. Moreover, we maximize the entropy of the plan to improve exploration. 
The augmented objective function is formulated as follows:

\begin{equation}
    \begin{aligned}
  \max_{\psi,\phi} \sum \limits_{t = 1}^T \mathbb{E}_{({\bf s}_t, {\bf p}_t,{\bf a}_t) \sim {\rho}_{(\kappa,\pi)}} \left[r_t ({\bf s}_t,{\bf p}_t, {\bf a}_t) + \alpha {\cal H}(\kappa_\psi (\cdot | {\bf s}_t))\right], 
    \end{aligned}
\end{equation}
where $\alpha$ is the temperature and ${\rho}_{(\kappa,\pi)}$ is a trajectory distribution under $\kappa_{\psi}( {\bf p}_t|{\bf s}_t)$ and $\pi_{\phi}( {\bf a}_t|{\bf p}_t)$.

\subsection{Learning Planner and Critic }

Unlike conventional RL algorithms, the critic $Q_{\theta}$ in our framework evaluates the plan ${\bf p}_t$ instead of the action ${\bf a}_t$ since learning a low-dimensional plan in an I2IT problem is easier and more effective. 
Specifically, the low-dimensional plan is concatenated into the downsampled vector of the critic and outputs the soft Q function $Q_{\theta}({\bf s}_t, {\bf p}_t)$, which is an estimation of the current state plan value, as shown in Fig.~\ref{ovspac}.

When the critic is used to evaluate the planner, rewards and soft Q values are used to iteratively guide the stochastic meta-policy improvement. 
In the evaluation step, by following SAC \cite{haarnoja2018soft}, \sw~learns $\kappa_\psi$ (planner) and fits parametric Q-function $Q_{\theta}({\bf s}_t,{\bf p}_t)$ (critic) using transitions sampled from the replay pool $\mathcal{D}$ by minimizing the soft Bellman residual:

\vspace{-1em}
\begin{equation*}
    \begin{aligned}
    &J_Q(\theta) = 
    \mathbb{E}_{({\bf s}_t,{\bf p}_t)\sim  \mathcal{D}} \left[\frac{1}{2} \Big(Q_{\theta}({\bf s}_t, {\bf p}_t) - \big(r_t + \gamma \mathbb{E}_{{\bf s}_{t+1}}\left[V_{\bar{\theta}}({\bf s}_{t+1})\right]\big)\Big)^2\right],
    \end{aligned}
\end{equation*}
where $V_{\bar{\theta}}({\bf s}_{t}) = \mathbb{E}_{{\bf p}_{t}\sim \kappa_\psi}[Q_{\bar{\theta}}({\bf s}_{t}, {\bf p}_{t}) - \alpha \log \kappa_{\psi} ({\bf p}_{t}|{\bf s}_{t})]$. We use a target network $Q_{\bar{\theta}}$ to stabilize training, whose parameters $\bar{\theta}$ are obtained by an exponentially moving average of parameters of the critic network \cite{lillicrap2015continuous}: $\bar{\theta} \rightarrow \tau \theta + (1-\tau)\bar{\theta}$. Hyper-parameter $\tau\in [0,1]$. To optimize $J_Q(\theta)$,  we can do the stochastic gradient descent with respect to parameters $\theta$ as follows,

\vspace{-1em}
\begin{equation}
    \begin{aligned}
    \theta = \theta &- \eta_Q \triangledown_{\theta} Q_{\theta}({\bf s}_t, {\bf p}_t)\Big(Q_{\theta}({\bf s}_t, {\bf p}_t) - r_t \\ 
    &- \gamma \left[Q_{\bar{\theta}}({\bf s}_{t+1}, {\bf p}_{t+1}) - \alpha \log \kappa_{\psi} ({\bf p}_{t+1}|{\bf s}_{t+1})\right]\Big).
    \end{aligned}
\label{eq:update_theta}
\end{equation}
Since the critic works on the planner, the optimization procedure will also influence the planner's decisions.
Following \cite{haarnoja2018soft}, we can use the following objective to minimize the KL divergence between the policy and a Boltzmann distribution induced by the Q-function,

\vspace{-1em}
\begin{equation*}
    \begin{aligned}
    J_{\kappa} (\psi) =& \mathbb{E}_{ {\bf s}_t\sim  \mathcal{D}}\big[ \mathbb{E}_{{\bf p}_t\sim  \kappa_{\psi}} \left[\alpha \log (\kappa_{\psi}({\bf p}_t| {\bf s}_t))-Q_\theta({\bf s}_t,{\bf p}_t)\right]\big]\\
    =&\mathbb{E}_{ {\bf s}_t\sim  \mathcal{D}, {\bf \epsilon}_t\sim  \mathcal{N} (\mu,\sigma) } \big[\alpha \log (\kappa_{\psi}( f_\psi({\bf \epsilon}_t,{\bf s}_t)|{\bf s}_t))
    - Q_\theta({\bf s}_t,f_\psi({\bf \epsilon}_t,{\bf s}_t))\big].
    \end{aligned}
\end{equation*}
The last equation holds because ${\bf p}_t$ can be evaluated by $f_\psi({\bf \epsilon}_t,{\bf s}_t)$ as we discussed before. It should be mentioned that hyperparameter $\alpha$ can be automatically adjusted by using the method proposed in~\cite{haarnoja2018soft}. Then we can apply the stochastic gradient method to optimize parameters as follows,

\vspace{-1em}
\begin{equation}
    \begin{aligned}
    \psi = &\psi -\eta_\psi\Big(\triangledown_\psi \alpha \log(\kappa_\psi({\bf p}_t|{\bf s}_t)) + \\
    &\big(\triangledown_{{\bf p}_t}\alpha \log(\kappa_{\psi}({\bf p}_t|{\bf s}_t))-\triangledown_{{\bf p}_t}Q_{\theta}({\bf s}_t,{\bf p}_t)\big) \triangledown_\psi f_\psi (\epsilon_t,{\bf s}_t)\Big).
    \end{aligned}
\label{eq:update_psi}
\end{equation}

The derivation for the case of the critic evaluating the actor can be found in \ref{app_critic_actor}. 
Besides, our experimental results 
also show that the critic evaluates actor’s action results in an inferior performance to that the critic
evaluates the planner.

\subsection{Task Specific Auxiliary Learning}


\begin{algorithm}[t]
    \caption{Learning Planner-Actor-Critic} \label{Alg0}
    \SetAlgoLined

    \KwIn{$I_F$, $I_M$, $U_F$, $U_M$, replay pool $\mathcal{D}$}
    
    
    \textbf{Init:} $\psi$, $\phi$, $\theta$, $\bar{\theta}$, $\mathcal{D}$ and environment $\cal E$
    
    \For{each iteration}{
    \For{each environment step}{
    ${\bf p}_{t} \sim \kappa_\psi({\bf p}_{t} | {\bf s}_{t})$,
    ${\bf a}_{t} \sim \pi_\phi({\bf a}_{t} | {\bf p}_{t})$
    
    ${\bf s}_{t+1}, {r_t} \sim \mathcal{U}({\bf s}_{t+1}|{\bf s}_{t},  {\bf p}_{t}, {\bf a}_{t})$
    
    $\mathcal{D} = \mathcal{D} \cup \{({\bf s}_{t}, {\bf p}_{t}, {\bf a}_{t}, r_t, s_{t+1})\}$
    }
    
    \For{each gradient step}{
    Sample from $\mathcal{D}$
    
    Update $\theta$, $\psi$, $\phi$ with Eq. (\ref{eq:update_theta}), Eq. 
    (\ref{eq:update_psi}), Eq. (\ref{eq:update_double}) 
    }
    }
\end{algorithm}


Following our meta policy (\(\kappa_{\psi}, \pi_{\phi}\)), the framework derives the executable action \({\bf a}_t\). To enhance convergence and performance, we adopt auxiliary learning for the planner and actor, tailored to specific tasks. 
This approach is highly adaptable and capable of integrating various advanced losses and techniques. 

For instance, in face inpainting tasks, we focus on reconstructing the predicted faces to match the original ones while also synthesizing more realistic images. This is achieved by employing a discriminator on predicted images with an adversarial loss. The nature of \({\bf a}_t\) is also task-dependent: 
for tasks aiming at realistic image generation, such as face inpainting or neural style transfer, \({\bf a}_t\) could directly be the target image \({\bf y}\). Detailed explanations and examples of these applications are provided in the experimental sections.
We elaborate on the auxiliary learning process using face inpainting tasks as an example. 
Concretely, the empirical objective of the reconstruction part in our framework is:

\vspace{-1em}
\begin{equation}
\centering
\label{rec_loss}
    {\cal L}_{rec}=\mathbb{E}_{{\bf s}_t,{\bf y} \sim {\cal D}}[\left \| {\cal T}({\bf s}_t)-{\bf y}\right \|_d],
\end{equation}
where $\cal D$ is a replay pool, $\left \| \cdot \right \|_d$ denotes some distance measure, such as $L_1$ or $L_2$. 
By adding a discriminator $D$ to predicted images, the adversarial loss is defined as:

\vspace{-1em}
\begin{equation}
\centering
\label{adv_loss}
    {\cal L}_{adv}= \mathbb{E}_{{\bf s}_t,{\bf y} \sim {\cal D}}[\log (D({\bf y})) + \log (1-D({\cal T}({\bf s}_t)))].
\end{equation}

In this example, the final auxiliary learning objective can be expressed as
\begin{equation}
\centering
\label{aux_loss}
    {\cal J}_{Aux} = {\lambda}_{rec} {\cal L}_{rec} + {\lambda}_{adv} {\cal L}_{adv},
\end{equation}
where ${\lambda}_{rec}$ and ${\lambda}_{adv}$ are the weight terms for reconstruction and adversarial learning.  
Finally, we can update $\psi$ and $\phi$  from the planner and the actor by performing the following steps: 
 \begin{equation}
    \begin{aligned}
    \psi = \psi-\eta\triangledown_\psi {\cal J}_{Aux}  (\psi, \phi), \ \ \ \phi = \phi-\eta\triangledown_\phi {\cal J}_{Aux}  (\psi, \phi).
    \end{aligned}
    \label{eq:update_double}
\end{equation}

Note that the additional auxiliary learning may introduce new parameters to learn, such as the discriminator $D$ in the above example. Since our goal is to learn the planner and the actor, which are the only components used in testing, we omit those additional notions for simplicity. More concrete examples of auxiliary learning are introduced in the experiment sections for different I2IT applications.
The pseudo-code of optimizing \sw~is described in Algorithm \ref{Alg0}. All parameters of \sw~are optimized based on the samples from replay pool $\cal D$. 




\subsection{Environment Settings in Practice}


In the given \sw~framework, environment designs are tailored for various applications, with detailed guidance in each section. This section outlines general principles for selecting rewards, focusing on the critic's role in evaluating plans rather than actions. The plans, being a subset of potential actions, serve as high-level instructions for the actor to create specific actions. Evaluation measures for structural or global image information, such as SSIM~\cite{wang2004image} or the DICE score \cite{dice1945measures}, are proposed as rewards for assessing these plans. 
However, the choice of reward remains flexible and should be empirically tested in the context of individual applications.

\vspace{-1em}

 \begin{figure}[t]
    \centering
    \includegraphics[scale=0.32]{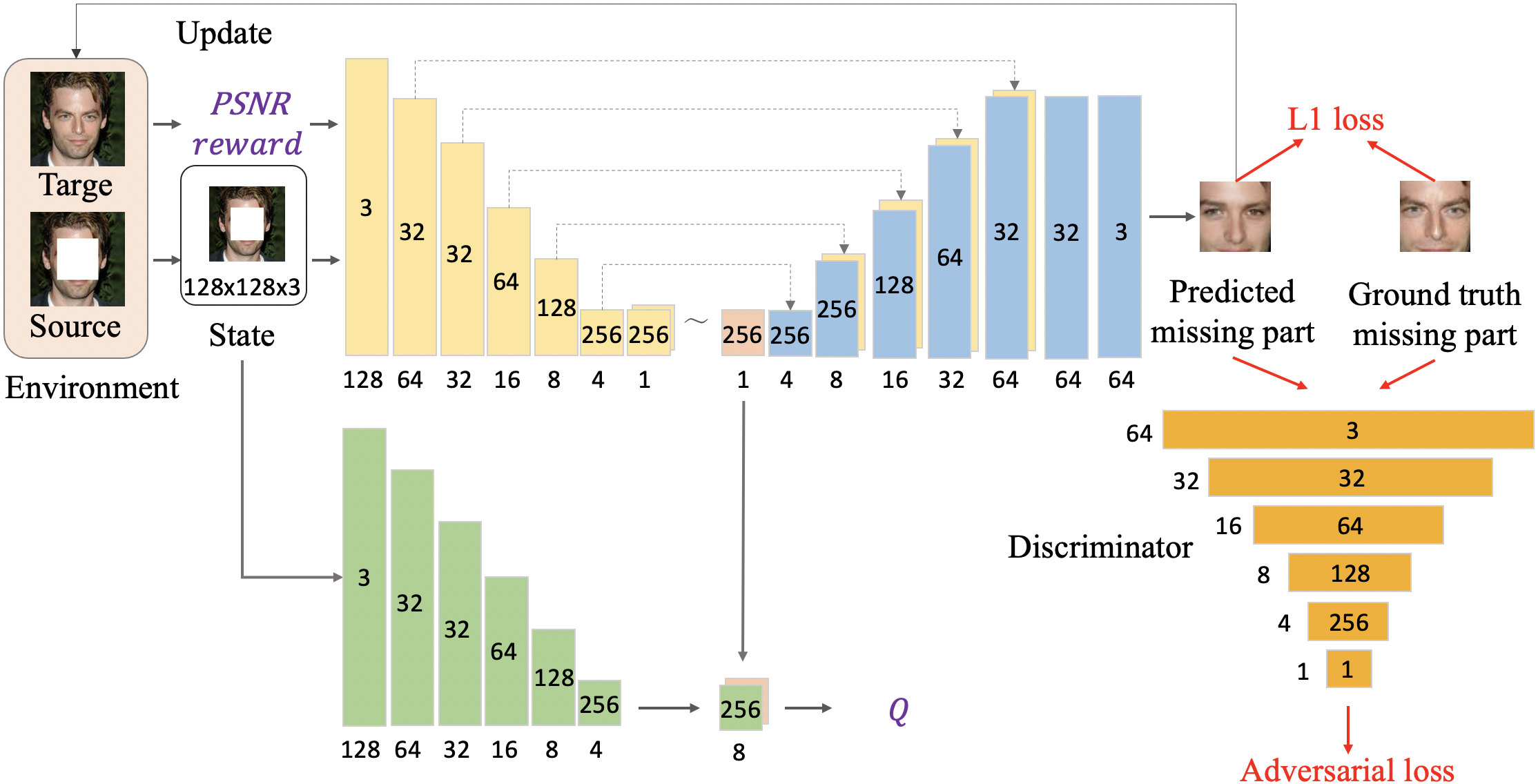}
    \caption{The network architecture of \sw~for face inpainting. Each rectangle represents a 2D image (or feature map), the number of channels is shown inside the rectangle, and the responding resolution is printed underneath (or on the left for discriminator).}
    \label{netfaceip}
\end{figure}



\section{Applications on Computer vision}
\label{app_cv}

\subsection{Face Inpainting}
\label{exp_fi}

In this section, we apply our \sw~framework to the face inpainting task, which aims to fill in a cropped region in the central area of a face with synthesized contents that are both semantically consistent with the original face and visually realistic.

\subsubsection{\sw~Setting}

For the state, we use the original image with a missing region (center cropped) as the initial state, and the next state is obtained by adding the new predicted image to the missing region. 
We use the peak signal-to-noise ratio (PSNR) as the reward. We apply the $L_1$ loss with an adversarial loss for the auxiliary learning, which tries to make the predicted image more realistic and closer to the ground truth image. 
The $\lambda_{rec}$ and $\lambda_{adv}$ in Eq.~\ref{aux_loss} are set to 1.0 and 0.02, respectively.






The network architecture for face inpainting is shown in Fig.~\ref{netfaceip}.
For the planner-actor, we use a similar architecture with context-encoder \cite{pathak2016context} except for the skip connections and the stochastic sampling operation in the planner. 
We use the same network structure for all the types of discriminators except minor changes for different GANs. Specifically, for WGAN-GP, the sigmoid function is removed from the final output layer. A spectral normalization is added to each layer of the discriminator of SNGAN~\cite{miyato2018spectral}. 
Moreover, the convolution layers of the planner, critic, and discriminator use $4 \times 4$ kernels, and the downsampling is performed by convolution with a stride of 2. 
In this application, the latent action dimension is set to 256. 




    


\subsubsection{Experiment}

We use the Celeba-HQ dataset in this task, which includes $28,000$ images for training and $2,000$ images for testing. 
All images have a cropped region of 64 $\times$ 64 pixels in the center. 
We compare our method 
with several recent face inpainting methods, including CE \cite{pathak2016context}, CA \cite{yu2018generative}, PEN \cite{zeng2019learning}, PIC \cite{zheng2019pluralistic}, RN \cite{yu2020region}, Shift-Net \cite{yan2018shift}, ILO \cite{daras2021intermediate}, and Palette \cite{saharia2022palette}. Following the previous work \cite{pathak2016context,yu2018generative,zheng2019pluralistic}, we use PSNR and SSIM as the evaluation metrics.

\begin{figure*}[t]
    \centering
    \includegraphics[scale=0.31]{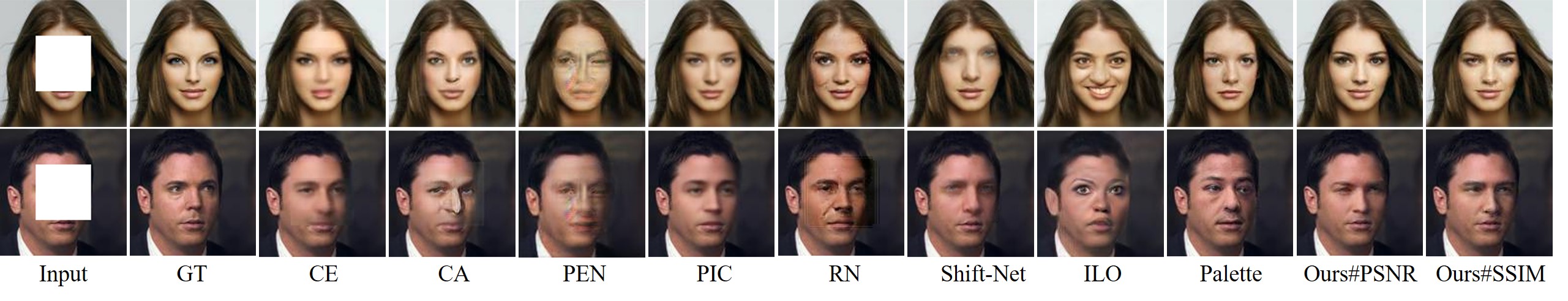}
    \caption{Visual comparison of different face inpainting methods. GT means ground truth. \sw~uses SNGAN for auxiliary learning. $\#$ indicates what reward is used for RL training. Our results have good visual quality even for a large pose face.}
    \label{fi-compare}
\end{figure*}

\noindent \textbf{Results and Analysis.} The qualitative results produced by our framework and existing state-of-the-art methods are shown in Fig.~\ref{fi-compare}. We can see easily that the \sw~gives obvious visual improvement for synthesizing realistic faces. 
The \sw~results are very reasonable, and the generated faces are sharper and more natural. 
This may be attributed to the high-level latent plan ${\bf p}_t$, which focuses on learning the global semantic structure and then directs the actor with auxiliary learning to further improve the local details of a generated image. 
We can also see that the synthesized images of the \sw~can have very different appearances from the ground truth, which indicates that, although our training is based on paired images, the \sw~can successfully explore and exploit data for producing diverse results.

The quantitative comparison is shown in Table~\ref{tab:psnr}. We can see that our method achieves the best PSNR and SSIM scores when compared with the existing state-of-the-art methods. As we mentioned before, the reward function in our RL framework is very flexible. Both PSNR- and SSIM-based rewards are suitable for face inpainting with the \sw~framework.

\begin{table}
\centering
\begin{tabular}{lcccc}
\toprule
\multicolumn{1}{l}{Method} & PSNR $\uparrow$ & SSIM $\uparrow$ & LPIPS $\downarrow$ & FID $\downarrow$ \\ \midrule
CE \cite{pathak2016context}                         & 25.764 & 0.850  &0.0955  &14.454   \\
CA \cite{yu2018generative}                  & 24.556  & 0.840   &0.0715   &9.950   \\ 
PIC \cite{zheng2019pluralistic}                      & 26.703    & 0.870  &0.0844  &12.470 \\
PEN \cite{zeng2019learning}                      & 23.196    & 0.634  &0.1342  &35.422  \\ 
RN \cite{yu2020region}                      & 25.123    & 0.835 &0.0698  &7.388 \\ 
Shift-Net \cite{yan2018shift}                              & 26.476    & 0.851  &0.0703  &7.597 \\
ILO \cite{daras2021intermediate}                                        & 22.709    & 0.783  &0.0958 &13.122  \\
Palette  \cite{saharia2022palette}                                   & 24.926    & 0.850  &0.0567  &4.909  \\
\midrule
Ours(PSNR)                  & 27.351   & 0.897  &0.0439  &\textbf{4.697}  \\
\textbf{Ours(SSIM)}           & \textbf{27.598}    & \textbf{0.899}  &\textbf{0.0433}  &4.917  \\
\bottomrule
\end{tabular}
\captionof{table}{Quantitative results of all methods on Celeba-HQ. We use SNGAN + PSNR and SNGAN + SSIM as rewards respectively.}
\label{tab:psnr}
\end{table}

\begin{table}[t]
\centering
\begin{tabular}{lcc}
\toprule
\multicolumn{1}{l}{Method} & PSNR $\uparrow$  & SSIM $\uparrow$  \\ \midrule
PA + SNGAN         & 26.884   & 0.871  \\ 
Ours (+ WGAN-GP)        &27.091   & 0.875 \\
Ours (+ RaGAN)        & 27.080    & 0.873\\
\textbf{Ours (+ SNGAN)}                        & \textbf{27.176}   & \textbf{0.882}   \\
\bottomrule
\end{tabular}
\captionof{table}{Ablation study of our \sw~framework on Celeba-HQ testing dataset (all trained with the PSNR reward).}
\label{tab:fi-gans}
\vspace{-1em}
\end{table}

\begin{table*}[t]
\renewcommand\arraystretch{1.1}
\centering
\scalebox{0.64}{
\begin{tabular}{lcccccccccccc}

\toprule
\multirow{3}{*}{Method} & \multicolumn{3}{c}{Facades label$\rightarrow$image}   &  \multicolumn{3}{c}{Cityscapes image$\rightarrow$label}   &
\multicolumn{3}{c}{Cityscapes label$\rightarrow$image}   &\multicolumn{3}{c}{Edges$\rightarrow$shoes} \\ 
 & PSNR  & SSIM  & LPIPS $\downarrow$  & PSNR & SSIM & LPIPS $\downarrow$ & PSNR  & SSIM & LPIPS $\downarrow$ & PSNR & SSIM & LPIPS $\downarrow$ \\ \midrule
 
pix2pix         & 12.290   & 0.225 & 0.438 & 15.891   & 0.457 & 0.287  &15.193  & 0.279 & 0.379 & 15.812   & 0.625 & 0.279  \\

PAN   & 12.779     & 0.249  & 0.387  & 16.317     & 0.566    & 0.228 & 16.408   & 0.391  &0.346  &16.097 &0.658 & 0.228\\

pix2pixHD   & 12.357     & 0.162  & 0.336  & 17.606     & 0.581    & 0.204 & 15.619   & 0.361  & \textbf{0.319}  &17.110 &0.686 & 0.220\\

DRPAN   & 13.101     & 0.276  & 0.354  & 17.724     & 0.633    & 0.214 & 16.673   & 0.403  &0.343  &17.524 &\textbf{0.713} & 0.221\\

CHAN   & 13.137    & 0.231   & 0.402  & 17.459    & 0.641   & 0.222 & 16.739    & 0.401   & 0.373  & 18.065    & 0.692   & 0.236\\

\midrule
Ours-PPO   & 13.163     & \textbf{0.308}  & 0.366  & 17.168     & 0.616    & 0.221 & 16.685   & 0.410  &0.362  &16.914 &0.695 & 0.225\\
\textbf{Ours}   & \textbf{13.178}   & 0.296  & \textbf{0.324} & \textbf{17.969}   & \textbf{0.659}  & \textbf{0.203}  & \textbf{16.848}  & \textbf{0.412}  & 0.337  & \textbf{18.178}   & 0.698 & \textbf{0.215} \\
\bottomrule
\end{tabular}}
\captionof{table}{Quantitative results of our \sw~and other methods over all datasets. $\downarrow$ means lower is better, Ours-PPO means our \sw~using PPO.}
\label{tab:pix2pix}
\end{table*}

\begin{table}[t]
\setlength{\abovecaptionskip}{0.15cm}
\setlength{\belowcaptionskip}{-0.2cm}
\centering
\scalebox{1.0}{
\begin{tabular}{lccccc}
\toprule
Method     & pix2pixHD  & DRPAN & CHAN  & Ours \\ \midrule
$\#$Params & 45.874M    & 11.378M  & 59.971M  & \textbf{9.730M} \\
$\#$FLOPs & 10.340G    & 14.208G  & 19.743G  & \textbf{3.519G}   \\ \bottomrule
\end{tabular}}
\caption{Comparison of the number of parameters and FLOPs (floating point operations, which represent the computational complexity of the model). 
}
\label{tab:params}
\end{table}

\begin{figure*}[t]
    \centering
    \includegraphics[scale=0.35]{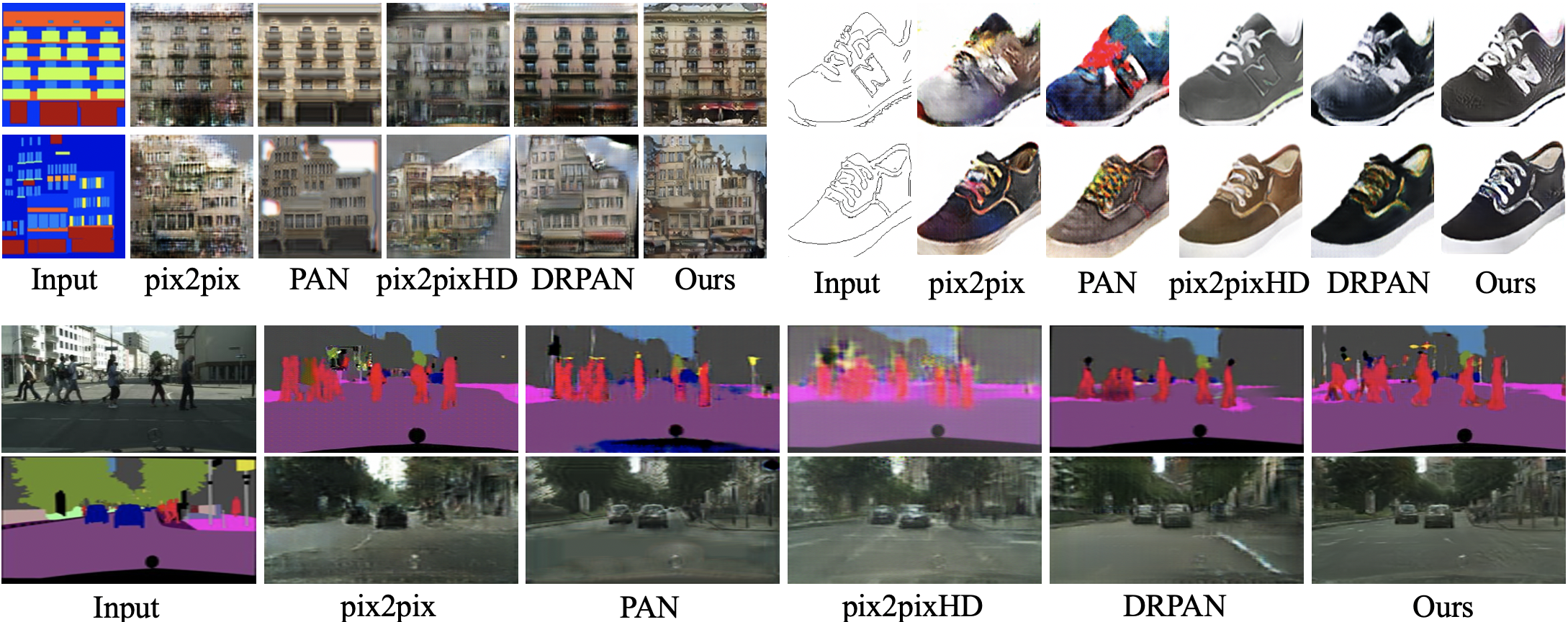}
    \caption{Visual comparison of our \sw~with pix2pix, PAN, pix2pixHD, and DRPAN over photo translation tasks.}
    \label{l2i}
    
\end{figure*}

\begin{figure*}[t]
\centering

\subfigure{
\includegraphics[width=0.24\textwidth]{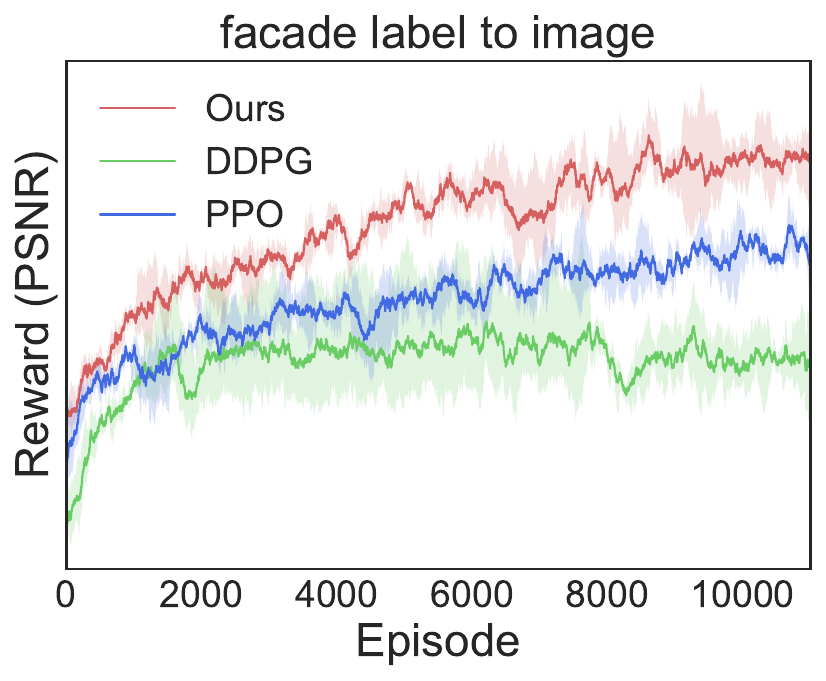}
\includegraphics[width=0.24\textwidth]{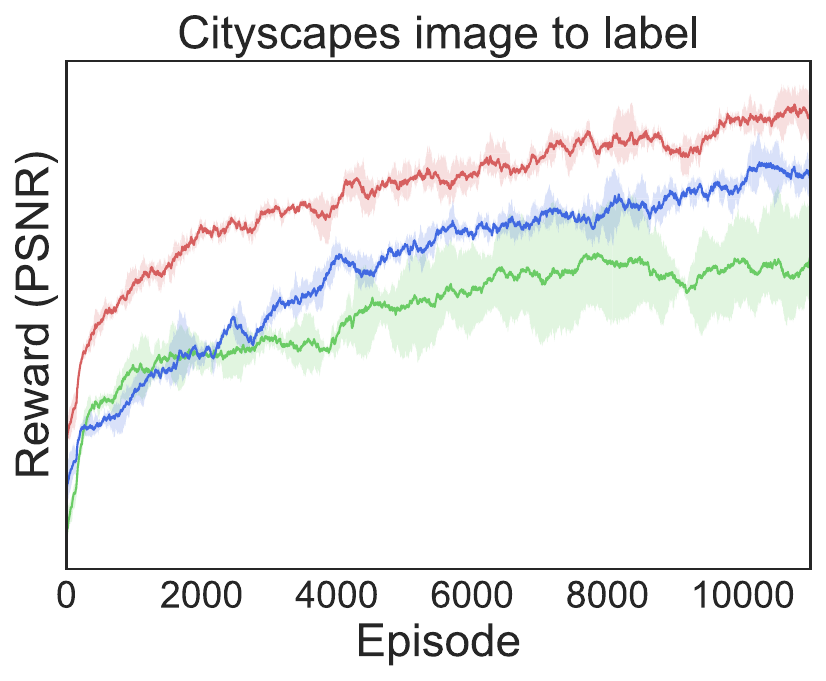}
 \includegraphics[width=0.24\textwidth]{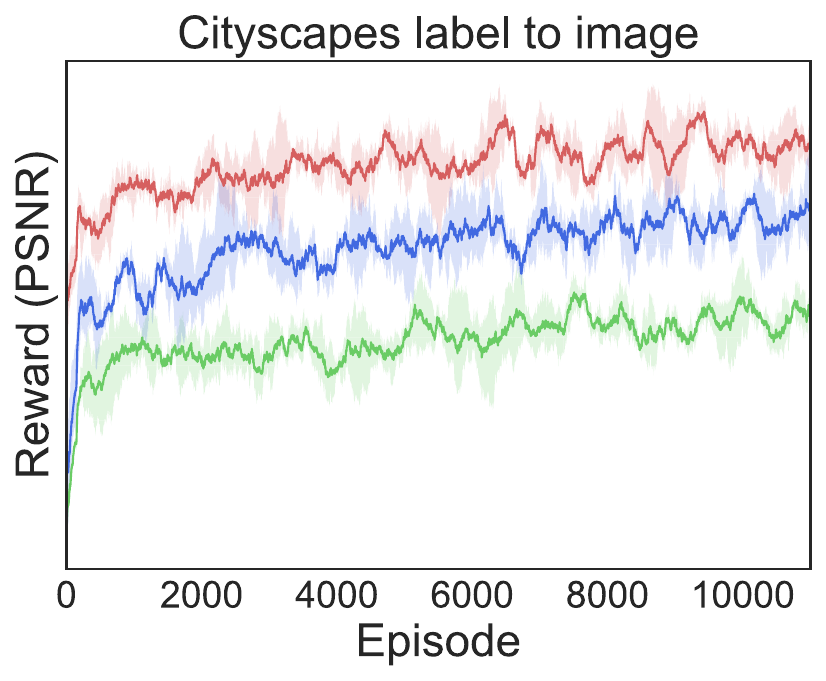}
 \includegraphics[width=0.24\textwidth]{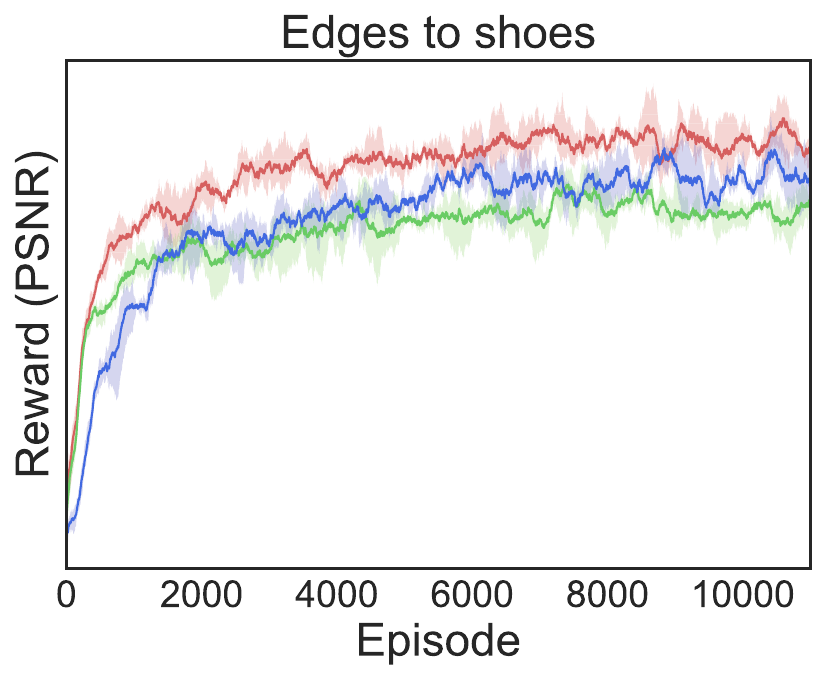}
}
\vspace{-1em}

\caption{Learning curves on different I2IT tasks. \sw~performs consistently better than other modified RL algorithms.}
\label{curves}
\vspace{-0.7em}
\end{figure*}

\noindent \textbf{Ablation Study.}
To illustrate the stability of training GANs in our framework, we jointly use $L_1$ and several advanced GAN losses, i.e., WGAN-GP \cite{gulrajani2017improved}, RaGAN \cite{jolicoeur2018relativistic},
and SNGAN \cite{miyato2018spectral} for auxiliary learning. 
We also separately train a planner-actor (PA) model by jointly optimizing the $L_1$ and the SNGAN loss. 
The results are shown in Table~\ref{tab:fi-gans}, which indicates that the \sw~framework with different GANs is stable and significantly improves the performance of training the planner-actor with SNGAN alone, further demonstrating the power of the \sw~framework.

\vspace{-1em}
\subsection{Realistic Photo Translation}
\label{exp_rpt}

In this section, we evaluate our \sw~framework on the general
realistic photo translation task.

\subsubsection{\sw~Setting} For realistic photo translation, we use the source image as the initial state. The next state is obtained by warping the generated image to the source image. 
We also let the action as the predicted image directly and use the same auxiliary learning settings and network structure as in the face inpainting experiment with the PSNR reward and the SNGAN loss (See Section \ref{exp_fi} for more details).

\subsubsection{Experiment}


We use three realistic photo translation tasks to evaluate our framework, 
(1) segmentation {\em labels$\rightarrow$images} with CMP Facades dataset \cite{tylevcek2013spatial},
(2) segmentation {\em labels$\rightarrow$images} and {\em images$\rightarrow$labels}  with Cityscapes dataset \cite{Cordts2016Cityscapes},
(3) {\em edges$\rightarrow$shoes} with Edge-shoes dataset \cite{yu2014fine}.

We compare our framework with existing methods, pix2pix \cite{isola2017image}, PAN \cite{wang2018perceptual}, and the methods designed for high-quality I2IT task, pix2pixHD~\cite{wang2018high}, DRPAN~\cite{wang2019drpan}, and CHAN~\cite{gao2021complementary}. 
Moreover, we replace MERL with PPO~\cite{schulman2017proximal}, denoted as Ours-PPO. We use PSNR, SSIM, and LPIPS \cite{zhang2018unreasonable} as the evaluation metrics.

\noindent \textbf{Results and Analysis.}
The quantitative results are shown in Table \ref{tab:pix2pix}. With a similar network structure, the proposed method significantly outperforms the pix2pix and PAN models on PSNR, SSIM, and LPIPS over all the datasets and tasks. 
Our method even achieves a comparable or better performance than the high-quality pix2pixHD and DRPAN models, which have much more complex architectures and training strategies. 
Moreover, using MERL instead of PPO obviously improves performance on most tasks. These experiments illustrate that the proposed \sw~framework is a robust and effective solution for I2IT.




More importantly, our model is much simpler, with the same architecture as pix2pix. The number of parameters and the computational complexity are shown in Table~\ref{tab:params}. We can see that 
the \sw~has much fewer parameters and lower computational complexity. 
We conclude that our model is lightweight, efficient, and effective.

The qualitative results of our \sw~with other I2IT methods on different tasks are shown in Fig.~\ref{l2i}. We can observe that pix2pix and PAN sometimes suffer from mode collapse and yield blurry outputs. The pix2pixHD is unstable on different datasets, especially on Facades and Cityscapes. 
The DRPAN is more likely to produce blurred artifacts in several parts of the predicted image on Cityscapes. 
In contrast, the \sw~produces more stable and realistic results. Using stochastic meta-policy and MERL helps explore more possible solutions so as to seek out the best generation strategy by trial-and-error in the training steps, leading to a more robust agent for different datasets and tasks.

%


\noindent \textbf{Evaluation of RL Algorithms.}
To demonstrate the effectiveness of stochastic meta policy and MERL,
we substitute the key components of \sw~with other structures or other state-of-the-art RL algorithms to test their importance. We use DDPG and PPO, respectively.
The learning curves of different variants on the four tasks are shown in Fig.~\ref{curves}, which indicates that,
by using the stochastic meta policy and the maximum entropy framework, the training process is significantly improved.

\begin{figure*}[t]
\centerline{
  \includegraphics[width=1\textwidth]{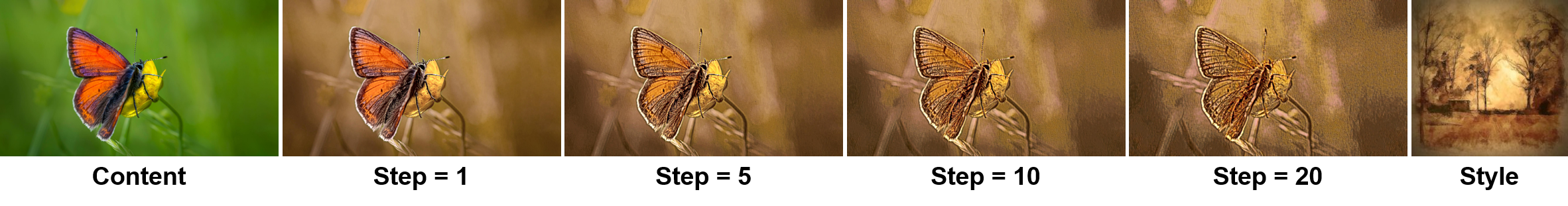}
  \vspace{-0.8em}  
}
\caption{ \small
Illustration of our step-wise style transfer process using the \sw~ framework. The content images are stylized stronger with the perdition steps smoothly. The model tends to preserve more details and structures of the content in the early steps and synthesize more style patterns in the later steps. Our framework allows users to control the stylization degree easily.
}
\label{introdemo} 
\vspace{-0.6em}
\end{figure*}

\subsection{Image Style Transfer }
\label{exp_nst}

Neural Style Transfer (NST)  refers to the generation of a pastiche image combining the semantic content of one image (the {\em content image}) and the visual style of the other (the {\em style image}) using a deep neural network. NST can be used to create a stylized non-photorealistic rendering of digital images with enriched expressiveness and artistic flavors. 


The one-step DL approach has an apparent limitation: it is hard to determine a proper level of style for different users since the ultimate metric of style transfer is too subjective. It has been observed that generated stylized images by the current NST methods tend to 
be under- or over-stylization~\cite{cheng2021style}. 
A remedy to under-stylization is to use the DL model multiple times, taking the output of the previous round as the input in the current round. However, this 
may suffer from the high computation cost due to the intrinsic complexity of one-step DL models. Other existing methods, like~\cite{gatys2015neural} and~\cite{huang2017arbitrary}, play a trade-off between content and style by adjusting hyper-parameters, but this approach is inefficient and hard to control.

Our \sw~framework provides a good solution for NST. It 
can be used to learn a lightweight NST model that is applied iteratively for NST. 
To preserve spatial structures of images,
the latent plans in our model are sampled from a 3D Gaussian distribution, which is estimated by the planner and forwarded to the actor, accompanied by style information
to generate intermediate images. 
Fig.~\ref{introdemo} shows some examples of our step-wise NST. We can see that our 
method tends to preserve more details and structures of the content image in early steps and synthesize more style patterns in later steps, resulting in a more flexible control of the stylization degree. Furthermore, our model is a lightweight and flexible NST model compared to existing methods, making it more efficient computationally. 
\begin{figure*}[t]
\centerline{
  \includegraphics[width=1\textwidth]{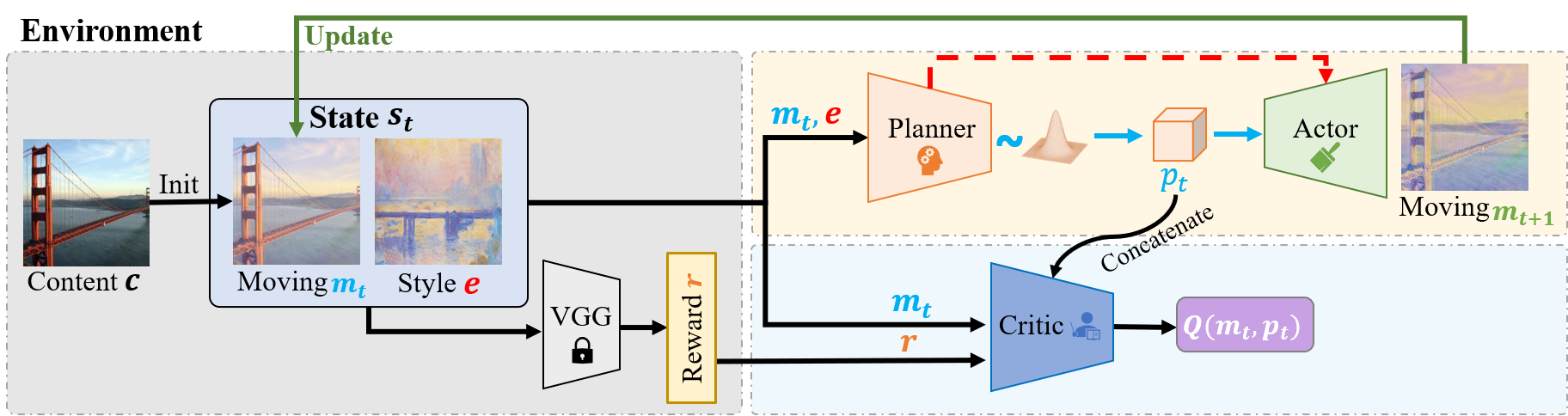}
  \vspace{-0.5em}
}
\caption{
Details of the \sw~ framework for the NST. The moving image is initialized with the content image. The plan is sampled from a 3D Gaussian distribution and is concatenated with the critic. The predicted moving image is generated by the actor. Note that the VGG networks are pre-trained and fixed for feature extraction during the training process.
}
\label{fig:pipline}
\vspace{-0.7em}
\end{figure*}

\subsubsection{\sw~Setting}
We set the moving image and style image as a whole as state ${\bf s}_t$, where the moving image is initialized by the content image. 
The moving image at time $t$ in state ${\bf s}_{t+1}$ is created by the actor and current state ${\bf s}_t$ and plan ${\bf p}_t$. 
The reward is obtained by measuring the difference between the current moving image in state ${\bf s}_t$ and the style image. 
The higher the difference is, the smaller the reward is. We use negative style loss as the reward. The style loss is defined later in this subsection. 


The detail of the framework is shown in Fig.~\ref{fig:pipline}. The planner is a neural network consisting of three convolutional blocks, two depthwise separable convolution blocks and two residual layers. In each convolutional block, after each convolutional layer, there is a ReLU layer. 
The planner estimates a 3D Gaussian distribution for sampling our latent plan with the size of 64$\times$64$\times$64, which is forwarded to the actor, accompanied by style information to generate the moving image.
The actor has two residual layers and three convolutional blocks. 
Our planner-actor is Fully Convolutional Network, which can process input images of any size. 
The critic consists of eight convolutional layers and one 1$\times$1 convolutional layer at the end. 
Since Johnson et al. \cite{johnson2016perceptual} conclude that using standard zero-padded convolutions in style transfer will lead to serious artifacts on the boundary of the generated image, we use reflection padding instead of zero padding for all the networks. More details are presented in the supplementary materials.

\textbf{Style Learning.}  
To make the moving image not deviate from the content image, the model trains the planner and the actor based on collected training data from the agent-environment interaction and changes dynamically in experience replay. 
More specifically, the planner and actor form a conditional generative process that translates state ${\bf s}_t$ to output moving image ${\bf m}_t$ at time $t$. 
Note that ${\bf s}_t$ is initialized to content image ${\bf c}$ and style image ${\bf e}$, and ${\bf s}_{t+1}$ 
is equivalent to the combination of ${\bf m}_t$ and ${\bf e}$.
Inspired by  \cite{wang2023microast}, we apply the content loss $\mathcal{L}^{CO}$ and style loss $\mathcal{L}^{ST}$ to optimize the model parameters of planner and actor. In addition, we introduce a new contrastive loss $\mathcal{L}^{CT}$ to enable the model to better distinguish different styles.
These losses can better measure perceptual and semantic differences between the moving image and input images. 

\textbf{Content Loss}. 
We evaluate the similarity between stylized image ${\bf m}_{t+1}$ and content image ${\bf c}$ by maximizing perceptual similarity using perceptual loss~\cite{johnson2016perceptual}, where content image ${\bf c}$ is iteratively updated by moving image ${\bf m}_t$.
Let $F^{(j)}$ denote the activation of the $j$-th layer, producing a feature map with dimensions $C^j \times H^j \times W^j$, where $C^j$, $H^j$, and $W^j$ represent the number of channels, height, and width of the feature map, respectively. The content loss $\mathcal{L}^{CO}$ is calculated by:
\begin{equation}
\mathcal{L}^{CO}({\bf m}_{t+1}, {\bf c})=\frac{\parallel F^{(j)}({\bf m}_{t+1})- F^{(j)}({\bf c})\parallel _{2}^{2}}{C^jH^jW^j}.
\end{equation}

\textbf{Style Loss}. 
The {style loss $\mathcal{L}^{ST}$} estimates the style deviations between the stylized image $ {\bf m}_{t+1} $ and style image $ {\bf e} $. Let $J$ represent the layer number of the network $F$.
It calculates statistical measures of $\mu$ and standard deviation $\sigma$ to penalize \( {\bf m}_{t+1} \), inspired by \cite{huang2017arbitrary}:
\begin{equation}
    \begin{aligned}
    \mathcal{L}^{ST}({\bf m}_{t+1}, {\bf e})= & \sum_{j=1}^J{\parallel}\mu (F^{(j)}({\bf m}_{t+1}))-\mu (F^{(j)}({\bf e}))\parallel _2^2\\
    & +\sum_{j=1}^J{\parallel}\sigma (F^{(j)}({\bf m}_t))-\sigma (F^{(j)}({\bf e}))\parallel _2^2.
    \end{aligned}
\label{eq:style_loss}
\end{equation}


\textbf{Hierarchical Style Representation Contrastive Loss (HSRCL):} Recent studies~\cite{chen2021artistic,wang2023microast} have demonstrated that a lightweight encoder struggles to effectively represent style images, and incorporating a contrastive learning loss can mitigate this issue. 
For instance, MicroAST~\cite{wang2023microast} employs a style signal contrastive learning loss to deal with this. 
However, our pilot study revealed that existing methods predominantly rely on deep-layer features for contrastive learning, overlooking the contributions of shallow-layer features to the overall stylistic representation.
To this end, we introduce a novel hierarchical style representation contrastive loss, which integrates comparative learning between deep and shallow feature representations, so as to enhance the stylistic representation. More specifically,when sampling a batch of data from the replay buffer $\cal D$, we construct both positive and negative sets for each sample's deep and shallow features.
And the feature contrastive
loss respectively computed from the deep features and shallow features are combined to create a hierarchical style contrastive loss function $\mathcal{L}^{CT}$, which is defined as:
\begin{equation}
    \mathcal{L}^{CT}=\sum_{i=1}^N{\sum_{k=1}^K{\frac{\parallel 
    \kappa_{\psi}({\bf m})^{(i,k)} - \kappa_{\psi}({\bf e})^{(i,k)}
    \parallel _2^2}{\sum_{j\ne i}^N{\parallel}
    \kappa_{\psi}({\bf m})^{(i,k)} - \kappa_{\psi}({\bf e})^{(j,k)}
    \parallel _2^2}}}.
\end{equation}
where $K$ represents the number of feature layers in the planner network, $N$ represents the batch size. The batch comprises $N$ states ${\bf S}=\{{\bf s}_1, {\bf s}_2, ..., {\bf s}_N\}$. Each state ${\bf s}_{i} \in {\bf S}$ consists of a moving image ${\bf m}^{(i)}$ and a style image ${\bf e}^{(i)}$. 
For each ${\bf m}$, we consider the style image ${\bf e}$ from ${\bf s}_{i}$ as a positive sample and the style images ${\bf e}$ from other ${\bf s}_{j }$ as negative samples.

\textbf{Uncertainty-aware Automatic Multi-task Learning (AML).} 
In style transfer, a common method to emhance the quality of style transfer involves quantifying the semantic similarity to the content image and the stylistic similarity to the style image through content and style loss functions, along with auxiliary loss functions, such as adversarial loss and total variation regularization. However, the weights for these loss functions are usually heuristically selected before training and remain unchanged throughout the training process, which is not sufficient enough to handle  images with different style and content.

To this end, as inspired by \cite{kendall2018multi}, we propose to use a multi-task learning framework that treats content learning, style learning, and contrastive learning as distinct but interconnected tasks. Using {\em homoscedastic uncertainty}, we dynamically adjust the loss weights of each task derived from a principled probabilistic model, achieving a balanced optimization objective that adapts throughout training. Unlike traditional methods requiring manual tuning of loss weights, our approach learns the relative importance of each task's loss function directly from the data. 
This not only simplifies the training process but also enables the dynamic modulation of content and style ratios to find the optimal solution.

Let $\lambda_c$, $\lambda_s$, $\lambda_{contrast}$ denote the loss weights for content loss, style loss, and contrastive loss, respectively. These weights can adapt based on homoscedastic uncertainty $\sigma_1^2$, $\sigma_2^2$, and $\sigma_3^2$, reflecting the noise level or task confidence. The loss weights are inversely proportional to the noise parameters: $\lambda_c=\frac{1}{\sigma_1^2}$, $\lambda_s=\frac{1}{\sigma_2^2}$, and $\lambda_{contrast}=\frac{1}{\sigma_3^2}$. The final loss is:
\begin{equation}
    \begin{aligned}
        \mathcal{L}_{final}(\psi, \phi, \sigma_1, \sigma_2, \sigma_3) = &
        \frac{1}{2\sigma_1^2} \mathcal{L}^{CO} + 
        \frac{1}{2\sigma_2^2} \mathcal{L}^{ST} + 
        \frac{1}{2\sigma_3^2} \mathcal{L}^{CT} + 
        \log(\sigma_1 \sigma_2 \sigma_3),
    \end{aligned}
\label{eq:awl}
\end{equation}
where $\log(\sigma_1 \sigma_2 \sigma_3)$ acts as a regularizer to prevent excessive increase in noise.  
Lastly, we employ a gradient descent method with learning rate $\eta$ to update the Planner and Actor parameters ($\psi$ and $\phi$) as well as $\sigma = (\sigma_1, \sigma_2, \sigma_3$):
\begin{equation}
    \psi \leftarrow \psi - \eta_\psi \triangledown_\psi \mathcal{L}_{final}, \quad
    \phi \leftarrow \phi - \eta_\phi \triangledown_\phi \mathcal{L}_{final},
    \label{eq:update_dl}
    \vspace{-1em}
\end{equation}

\begin{equation}
    \sigma \leftarrow \sigma - \eta_\sigma \triangledown_\sigma \mathcal{L}_{final}.
    \label{eq:update_dl2}
\end{equation}



\begin{figure*}[t]
\centerline{
  \includegraphics[width=1\textwidth]{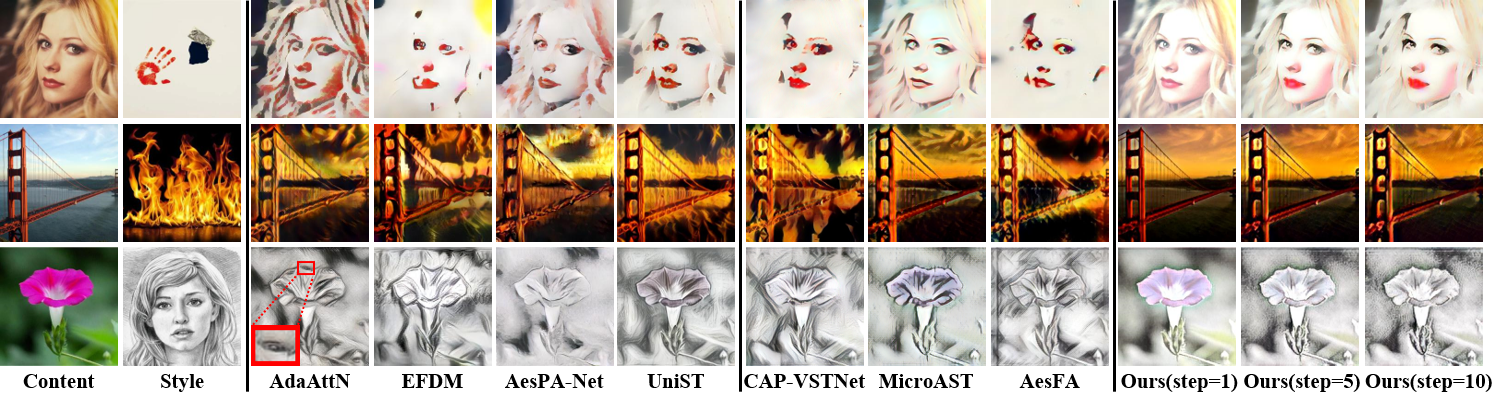}
  \vspace{-0.8em}
}
\caption{Qualitative Comparison in 256 pixel resolution. The first two columns show the content and style images, respectively. The subsequent four columns display the results from the current SOTA NST methods. The three columns immediately following showcase the results of the lightweight methods. Lastly, we present the step-wise stylization results generated by our method.
}
\label{fig:vision_compare}
\vspace{-0.8em}
\end{figure*}

\subsubsection{Experiment}
\noindent \textbf{Dataset}. 
Like most NST methods~\cite{deng2022stytr2,huang2017arbitrary,liu2021adaattn,park2019arbitrary,wang2023microast}, we utilize the MS-COCO dataset~\cite{lin2014microsoft} for content and the WikiArt dataset~\cite{phillips2011wiki} for style. During training, images are first scaled to 512$\times$512 pixels, then randomly cropped to 256$\times$256, while testing can handle any input size. Following MicroAST~\cite{wang2023microast}, we assess all algorithms across eight aspects: {\em visual effect, inference time, parameter count, GFLOP, content loss, style loss, SSIM~\cite{wang2004image}, and storage space}.

\noindent \textbf{Baselines}.
We compare our method with three light-weight NST methods: CAP-VSTNet~\cite{wen2023cap}, MicroAST~\cite{wang2023microast} and AesFA~\cite{kwon2023aesfa}, as well as four SOTA NST methods: AdaAttN~\cite{liu2021adaattn}, EFDM~\cite{zhang2022exact}, AesPA-Net~\cite{hong2023aespa} and UniST~\cite{gu2023two}.  All codes used in the experiment are sourced from public repositories, and we use the default settings provided.

\begin{table*}[t]
\centerline{
\resizebox{1.0\linewidth}{!}{
\setlength{\tabcolsep}{1.2mm}
\begin{tabular}{c|cc|ccclc|ccclc}
\hline
         & \multicolumn{2}{c|}{Model Complexity}      & \multicolumn{5}{c|}{256$\times$256 Pixel Resolution}         & \multicolumn{5}{c}{512$\times$512 Pixel Resolution}                       \\ \hline
Method        & \multicolumn{1}{c}{Params (1e6) $\downarrow$} & \multicolumn{1}{c|}{Storage (MB) $\downarrow$} & Content Loss 
 $\downarrow$    & SSIM $\uparrow $          & \multicolumn{1}{c|}{Style Loss $\downarrow$}      & \multicolumn{1}{c|}{Time(s) $\downarrow$} & Pref.(\%) $\uparrow$ & Content Loss $\downarrow$ & SSIM $\uparrow$ & \multicolumn{1}{c|}{Style Loss $\downarrow$} & \multicolumn{1}{c|}{Time(s) $\downarrow$} & Pref.(\%) $\uparrow$ \\ \hline
AdaAttN(2021)     &13.6299   &128.4020  &3.0668   &0.4987   &\multicolumn{1}{c|}{0.6027} &\multicolumn{1}{c|}{0.0117} &9.00   &2.4280   &0.5341  &\multicolumn{1}{c|}{0.5516}  &\multicolumn{1}{c|}{0.1032}  &11.00  \\
EFDM(2022)        &7.0110    &26.7000  &3.6671   &0.3165   &\multicolumn{1}{c|}{0.4233} &\multicolumn{1}{c|}{0.0073}  &4.67 &2.9439   &0.3788   &\multicolumn{1}{c|}{0.3268}  &\multicolumn{1}{c|}{0.0079} &6.67    \\
CAP-VSTNet(2023)  &4.0899    &15.6719  &3.5984   &0.4501   &\multicolumn{1}{c|}{\textbf{0.3151}} &\multicolumn{1}{c|}{0.0423} &7.33   &2.7459   &0.4864   &\multicolumn{1}{c|}{\textbf{0.2234}}  &\multicolumn{1}{c|}{0.1209} & 7.33 
 \\
AesPA-Net(2023)   &23.6737   &92.3340  &2.6822   &0.4504   &\multicolumn{1}{c|}{0.8266}  &\multicolumn{1}{c|}{0.3110} &6.00  &2.0412   &0.5195   &\multicolumn{1}{c|}{0.8756}  &\multicolumn{1}{c|}{0.4628}  &10.00   \\
UniST(2023)       &65.2545   &302.9424  &2.8888   &0.4305   &\multicolumn{1}{c|}{0.4137} &\multicolumn{1}{c|}{0.0295}  &8.33   &2.4080   &0.4567   &\multicolumn{1}{c|}{0.2952}  &\multicolumn{1}{c|}{0.0347} &8.00   \\
MicroAST(2023)    &0.4720    &1.8570  &2.6382   &0.4753   &\multicolumn{1}{c|}{0.6247}  &\multicolumn{1}{c|}{\textbf{0.0066}}  &9.00  &2.0349   &0.5034   &\multicolumn{1}{c|}{0.4960}  &\multicolumn{1}{c|}{\textbf{0.0069}} &8.33  
 \\
AesFA(2024)       &3.2208    &12.3100  &3.3734   &0.4115   &\multicolumn{1}{c|}{0.3945}  &\multicolumn{1}{c|}{0.0167}  &8.34  &2.7624   &0.4466   &\multicolumn{1}{c|}{0.3024}  &\multicolumn{1}{c|}{0.0187} &7.67   \\ \hline
Ours($1^{st}$)      &\textbf{0.3712}    &\textbf{1.4750}  &\textbf{1.1684}   &\textbf{0.6444}   &\multicolumn{1}{c|}{1.0487} &\multicolumn{1}{c|}{0.0094}  &\textbf{20.33}   &\textbf{0.9292}   &\textbf{0.6517}   &\multicolumn{1}{c|}{0.8927}  &\multicolumn{1}{c|}{0.0150}   &\textbf{19.00}   \\
Ours($5^{th}$)      &\textbf{0.3712}    &\textbf{1.4750}  &\textbf{2.1508}   &\textbf{0.5509}   &\multicolumn{1}{c|}{0.6974} &\multicolumn{1}{c|}{0.0336}  &\textbf{22.33}   &\textbf{1.6491}   &\textbf{0.5711}   &\multicolumn{1}{c|}{0.5528}  &\multicolumn{1}{c|}{0.0852}  &\textbf{13.67}  \\
Ours(${10}^{th}$)     &\textbf{0.3712}    &\textbf{1.4750}  &2.7518   &0.4898   &\multicolumn{1}{c|}{0.6209} &\multicolumn{1}{c|}{0.0631}   &4.67   &2.0871   &0.5191   &\multicolumn{1}{c|}{0.4892}  &\multicolumn{1}{c|}{0.1733}  &8.33   \\ \hline
\end{tabular}}
}
\caption{Quantitative Comparison of Model Complexity and Performance with Various AST Algorithms at Standard Resolutions. `Pref.' represents user preferences from our user study.  }
\label{tab:compare_256_512}
\end{table*}

\begin{table*}[thb]
\centerline{
\resizebox{1.0\linewidth}{!}{
\setlength{\tabcolsep}{1.2mm}
\begin{tabular}{c|cccc|cccc|cccc}
\hline
Resolution     & \multicolumn{4}{c|}{FHD (1K: $1920\times 1080$)}                      & \multicolumn{4}{c|}{QHD (2K: $2560 \times 1440$)}          & \multicolumn{4}{c}{UHD (4K: $3840\times 2160$)}                               \\ \hline
Method  &Content Loss$\downarrow$  &SSIM$\uparrow$  &\multicolumn{1}{c|}{Style Loss$\downarrow$} & Time(s)$\downarrow$  &Content Loss$\downarrow$  &SSIM$\uparrow$  &\multicolumn{1}{c|}{Style Loss$\downarrow$} & Time(s)$\downarrow$  &Content Loss$\downarrow$  &SSIM$\uparrow$  &\multicolumn{1}{c|}{Style Loss$\downarrow$} & Time(s)$\downarrow$ \\ 
\hline
EFDM(2022)       &2.3051   &0.4331     &\multicolumn{1}{c|}{0.2975}  &0.0085   &2.1531   &0.4381   &\multicolumn{1}{c|}{0.2889} &0.0104  &1.9927   &0.4345  &\multicolumn{1}{c|}{0.2948}  &0.0156  \\
CAP-VSTNet(2023) &2.0444   &0.5194     &\multicolumn{1}{c|}{\textbf{0.1814}} &1.0174  &1.9079 &0.5230 &\multicolumn{1}{c|}{\textbf{0.1640}} &1.8145   &1.8057 &0.5200 &\multicolumn{1}{c|}{\textbf{0.1518}}  &4.0979\\
MicroAST(2023)   &1.5372    &0.5243   &\multicolumn{1}{c|}{0.4101} &\textbf{0.0074}   &1.4366   &0.5233   &\multicolumn{1}{c|}{0.3749}  &\textbf{0.0090}   &1.3454   &0.5156  &\multicolumn{1}{c|}{0.3444} &\textbf{0.0109}  \\
AesFA(2024)      &2.1742    &0.4799   &\multicolumn{1}{c|}{0.2665}  &0.0191   &2.0330   &0.4834   &\multicolumn{1}{c|}{0.2500}  &0.0191   &1.9073   &0.4793  &\multicolumn{1}{c|}{0.2425} &0.0195  \\ \hline
Ours($1^{st}$)     &\textbf{0.7097}  &\textbf{0.6558}  &\multicolumn{1}{c|}{0.8216}  &0.0851   &\textbf{0.6572}   &\textbf{0.6528}   &\multicolumn{1}{c|}{0.7712}  &0.1473   &\textbf{0.6254}   &\textbf{0.6409}  &\multicolumn{1}{c|}{0.7207}  &0.3242 \\
Ours($5^{th}$)     &\textbf{1.2330}  &\textbf{0.5805}  &\multicolumn{1}{c|}{0.4726}  &0.6022  &\textbf{1.1390}   &\textbf{0.5776}   &\multicolumn{1}{c|}{0.4315}  &1.0586   &\textbf{1.0695}   &\textbf{0.5657}   &\multicolumn{1}{c|}{0.3898} &2.3688  \\
Ours(${10}^{th}$)    &1.5448  &\textbf{0.5361}  &\multicolumn{1}{c|}{0.4165}  &1.2480   &\textbf{1.4204}   &\textbf{0.5358}   &\multicolumn{1}{c|}{0.3766}  &2.2013   &\textbf{1.3200}   &\textbf{0.5259}   &\multicolumn{1}{c|}{0.3365}  &4.9304 \\ \hline
\end{tabular}}
}
\caption{Quantitative comparison with lightweight AST algorithms at different high resolutions, the best results are highlighted in bold. 
Among these lightweight methods, our method demonstrates competitive results. As the sequence increases in our method, the style loss gradually decreases while the semantic information is preserved to the greatest extent.}
\label{tab:compare_various_resolution}
\end{table*}

\noindent \textbf{Implementation Details}.
We use the Adam optimizer~\cite{kingma2014adam} with learning rate 2e-4, the batch size in the environment set to 1, and the batch size sampled from the replay buffer set to 8. All experiments are conducted on a single NVIDIA Tesla P100 (16GB) GPU.

\noindent \textbf{Qualitative Comparison}.
We visually compare our method with all baseline methods in Fig.~\ref{fig:vision_compare}. AdaAttN shows a repetitive stylistic pattern resembling the eyes (the third row), while EFDM and CAP-VSTNet lose a significant semantic and structural content (first and second rows). AesPA-Net produces inconsistent results, especially in the eye area (first row). UniST and MicroAST show insufficient stylization (third row), and AesFA has severe boundary artifacts (third row). In contrast, our approach generates a sequence of results with increasing stylization levels while maintaining coherent content structure. 
In the supplementary materials, we also compared our method with lightweight baseline methods at higher resolutions.

\noindent \textbf{Quantitative Results.}
Tables~\ref{tab:compare_256_512} and \ref{tab:compare_various_resolution} provide a comprehensive  comparison between our approach and baseline models, covering model complexity and performance metrics at both standard and high resolutions. Our method consistently achieves competitive scores in content loss, SSIM, style loss, and inference time, demonstrating its efficiency and effectiveness in producing outputs that balance stylistic expression with content preservation. As the sequence progresses, our method enhances stylistic richness while maintaining content fidelity. In terms of model complexity, our model is 20\% smaller than the smallest baseline model, achieving a more streamlined architecture. Additionally, it is the first AST method capable of controlling the degree of stylization on images ranging from 256 to 4K resolution.

\begin{figure*}[t]
\centerline{
  \includegraphics[width=\textwidth]{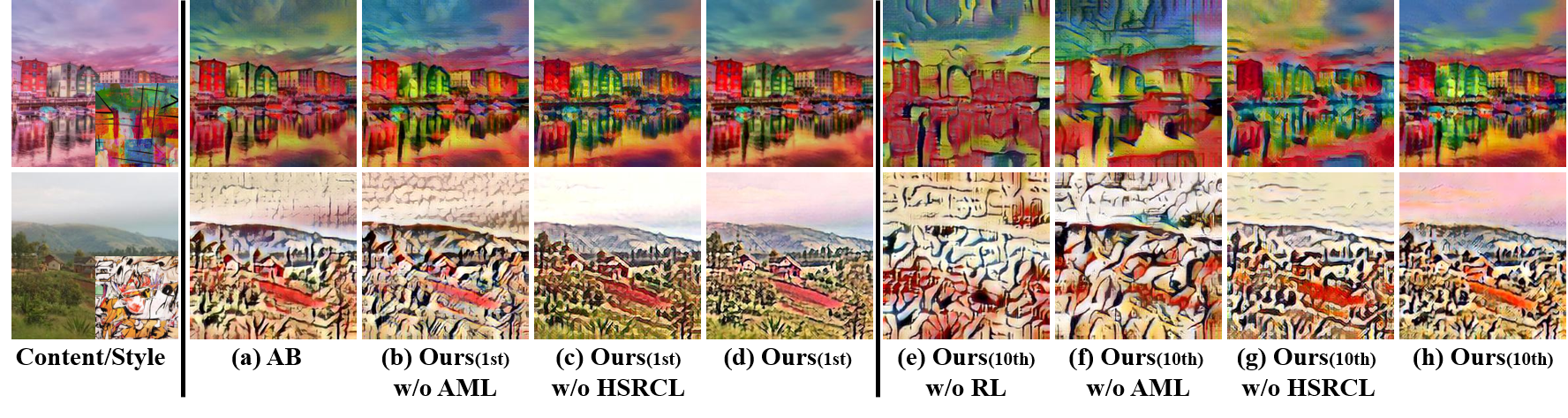}
  \vspace{-2mm}
}
\caption{
Ablation Study Results Comparing the Impact of RL, 
AML, and HSRCL vs. Style Signal Contrastive Loss on Style Transfer Performance. 
The visual comparison underscores the contributions of RL, AML, and HSRCL to the fidelity and stability of stylized results across sequences.
}
\label{fig:ablationstudy1}
\end{figure*}

\noindent \textbf{Ablation Study.}
(1)We first discussed the effectiveness of RL in style control.
In Fig.~\ref{fig:ablationstudy1},
without RL, the Actor-Builder (AB) in (a) initially preserves semantic information but introduces incorrect textures in the background (second row). At sequence 10, notable content information is lost. In contrast, our method in (d) produces smoother and clearer stylized images from the start, and stably maintains high-quality results throughout the sequence in (h) at sequence 10. This consistent performance highlights the significant enhancement of RL provides to DL-based AST models.

(2)We manually tuned the loss weights in our method, based on the settings of MicroAST and empirical adjustments. Specifically, we set the content loss weight $\lambda_{CO}=1$, the style loss weight $\lambda_{ST}=3$, and the HSRCL loss weight $\lambda_{CT}=3$, while keeping all other settings unchanged. 
As shown in Fig.~\ref{fig:ablationstudy1} , compared to the fixed loss weight method in (b,f), our approach using Automatic Multi-task Learning (AML) demonstrates superior content preservation in both sequence 1 in (d) and sequence 10 in (h). Our study indicates that AML significantly enhances model performance and accelerates network convergence.

(3)We investigated the effectiveness of HSRCL by comparing with the deep-feature based contrastive loss proposed in MicroAST~\cite{wang2023microast}. 
As shown in Fig.~\ref{fig:ablationstudy1}, for sequence 1, using only deep features for contrastive learning (see Fig. 12(c)) exhibits less of stylistic diversity as compared with the result in Fig. 12(d). Comparing with sequence 10 in Fig. 12(h), there is a noticeable decline in Fig. 12(g) in terms of content affinity due to incoherent stylistic expression. This experiment demonstrates that HSRCL significantly enhances the model's capacity in stylistic expression.

\noindent \textbf{User Study.}
Evaluating the results of style transfer is highly subjective. Therefore, we conducted user study on eight different methods. We recruited 30 participants representing a diverse range of ages, genders, and professional backgrounds. Each participant was randomly presented with 20 ballots: 10 at a $256\times 256$ resolution and $512 \times 512 $ resolution. Each ballot included the content image, the style image, and 10 randomly shuffled stylized results. Note that since our method produces sequential results, we present the outcomes at the first, fifth, and tenth sequences. 
We collected a total of 300 valid ballots, and the detailed results are shown in Table~\ref{tab:compare_256_512}. It is evident that the majority of users prefer the stylized results generated by our method.
In other words, although the assessment of stylized results is inherently subjective, our lightweight style transfer agent is designed to generate a diverse array of sequential outputs tailored to meet the varying preferences and requirements of different users.

\vspace{-0.5em}
\section{Discussion}
\label{discussion}

\subsection{General Framework}
\sw~is designed with a core focus on providing a solution that is not limited to a single task. Our framework features a highly modular design that allows it to adapt effortlessly to a variety of image translation tasks, including face inpainting, neural style transfer, and deformable image registration. In each application, the \sw~framework not only matches the performance of state-of-the-art methods but also surpasses them on certain key metrics. For instance, in the task of neural style transfer, our model excels in both the diversity and quality of style transfer while maintaining consistent content, outperforming existing technologies. And 
the flexibility and extensibility of the framework are evident in its ability to seamlessly integrate advanced auxiliary learning techniques such as Generative Adversarial Networks (GANs) and autoencoders. The incorporation of these technologies not only enhances the model's performance in image generation tasks but also provides tailored solutions for specific problems.

\subsection{Differences from Diffusion Models}
Diffusion models typically employ a multi-step denoising process to generate data, which involves incrementally introducing noise and then reversing the process to remove it. Each step builds upon the result of the previous one, culminating in the final output. While these models can produce high-quality images, they require the full execution of all steps, lacking utility in the intermediate stages.

In contrast, the \sw~framework allows for results at each step that possess a degree of practicality.
This framework utilizes reinforcement learning for decision-making, with each output being usable for further iterative refinement or as an intermediate result when needed. For instance, in style transfer tasks, the \sw~framework can generate results that progressively align with the target style at each step, permitting users to evaluate throughout the iterative process.

Furthermore, the step-wise practicality of the \sw~framework offers greater flexibility to users, enabling them to halt the iterative process at any step to obtain satisfactory outputs according to their specific requirements. This flexibility is not present in diffusion models, which often necessitate the completion of the entire denoising sequence to achieve satisfactory results.

\subsection{Limitations}
While our \sw~framework has achieved significant results in image-to-image translation tasks, we acknowledge that there are some limitations to this study. 
Specifically, in terms of model interpretability, although the framework can produce high-quality translation results, the decision-making process and internal mechanisms of the model remain somewhat opaque. For instance, when performing complex style transfer tasks, the criteria by which the model selects specific stylistic features are not clearly defined, which constrains users' understanding and trust in the model's behavior.

In addition, the contributions of the planner and critic modules have not been analyzed in isolation. Nevertheless, their importance is indirectly validated by our ablation experiments and training observations. Removing the RL component effectively eliminates the critic’s guidance, which leads to a substantial drop in performance (as shown in Fig.~\ref{fig:ablationstudy1}). On the other hand, if the planner is removed, the actor no longer receives low-dimensional semantic plans to guide action generation. In this case, the network cannot operate properly, as the planner serves as the encoder bridging the high-dimensional state space and the action space. This demonstrates that both the planner and critic are indispensable, although more fine-grained module-level ablations remain for future work.

Finally, the current framework employs a fixed maximum number of inference steps. Since there is no ground-truth target available during testing, automatic stopping based on reconstruction metrics (e.g., PSNR or SSIM) is not feasible. At present, inference proceeds until the predefined maximum step, which provides robustness but may reduce efficiency in some cases. Developing a learned adaptive stopping policy that does not rely on external targets is therefore an important direction for future refinement.

\vspace{-0.5em}

\section{Conclusion}
\label{conclusion}

In this paper, we propose a reinforcement learning-based framework, \sw, to handle the I2IT problem. Our \sw~framework is an off-policy planner-actor-critic model. It can efficiently learn good policies in spaces with high-dimensional continuous states and actions. The core component in \sw~is the proposed meta policy with a new component `plan', which is defined in latent subspace and can guide the actor to generate high-dimensional executable actions. 
To the best of our knowledge, we are the first to propose an RL framework for the I2IT problem. Experiments based on diverse applications demonstrate that this architecture achieves significant gains over existing SOTA methods.


The framework we propose has several potential limitations. One such limitation is that our framework is only capable of performing NST tasks on images. Video style transfer typically requires a higher degree of temporal consistency, whereas our RL-based framework primarily interacts with the current state, focusing more on the diversity of the results. This difference limits the applicability of our framework to video style transfer. Nonetheless, the main goal of the NST tasks demonstrated in this paper is to validate the effectiveness of our RL-based method in controlling stylistic elements and its superiority in achieving optimal NST quality. For the image-based NST model, this is sufficient to meet our needs. However, we recognize that by introducing temporal consistency components or specific loss functions, the current framework can be extended to support video style transfer tasks. Another potential limitation is that the number of steps of the testing process is a predefined hyper-parameter, which can be improved by learning from the model automatically.

In the future, we will try to address the aforementioned limitations of our proposed framework. 
We expect that the proposed architecture can potentially be extended to all I2IT tasks.


%

\vspace{-1em}

\section*{Acknowledgments}

This work was supported in part by the National Natural Science Foundation of China under Grants 42375148, Sichuan province Key Technology Research and Development project under Grant No.2024ZHCG0190, No. 2024ZHCG0176. CUIT Science and Technology Innovation Capacity Enhancement Program project under Grant KYQN202305

Xin Wang is supported by University at Albany, SUNY Start-up Grant.
\vspace{-1em}


\appendix

\section{Learning with Critic on Actor}
\label{app_critic_actor}

When the critic is used to evaluate the actor, the rewards and the soft Q values are used to guide the stochastic policy improvement iteratively, 
where the ${\bf a}_t$ is concatenated on the state ${\bf s}_t$ as the input of the critic. 
In evaluation step, follow SAC \cite{haarnoja2018soft}, \sw~learns the actor $\pi_\phi$ and fits the parametric Q-function $Q_{\theta}({\bf s}_t,{\bf a}_t)$ (critic) using transitions sampled from the replay
pool $\mathcal{D}$ by minimizing the soft Bellman residual,

\vspace{-1em}
\begin{equation}
    \begin{aligned}
    J_Q(\theta) = \mathbb{E}_{\mathcal{D}} \left[\frac{1}{2} \Big(Q_{\theta}({\bf s}_t, {\bf a}_t) -  \big(r_t + \gamma \mathbb{E}\left[V_{\bar{\theta}}({\bf s}_{t+1})\right]\big)\Big)^2\right],
    \end{aligned}
\end{equation}
where $V_{\bar{\theta}}({\bf s}_{t}) = \mathbb{E}_{{\bf a}_{t} \sim \pi_{\phi}} [Q_{\bar{\theta}}({\bf s}_{t}, {\bf a}_{t}) - \alpha \log \pi_{\phi} ({\bf a}_{t}|{\bf p}_{t})]$. $\gamma$ is the discount factor. We use a target network $Q_{\bar{\theta}}$ to stabilize training, whose parameters $\bar{\theta}$ are obtained by an exponentially moving average of parameters of the critic network \cite{lillicrap2015continuous}: $\bar{\theta} \rightarrow \tau \theta + (1-\tau)\bar{\theta}$. The hyper-parameter $\tau\in [0,1]$. To optimize the $J_Q(\theta)$,  we can do the stochastic gradient descent \cite{haarnoja2018soft} with respect to the parameters $\theta$ as follows,

\vspace{-1em}
\begin{equation}
    \begin{aligned}
    \theta = \theta - \eta_Q &\triangledown_{\theta} Q_{\theta}({\bf s}_t, {\bf a}_t)\Big(Q_{\theta}({\bf s}_t, {\bf a}_t) - r_t \\
    &- \gamma \left[Q_{\bar{\theta}}({\bf s}_{t+1}, {\bf a}_{t+1}) - \alpha \log \pi_{\phi} ({\bf a}_{t+1}|{\bf p}_{t+1})\right]\Big).
    \end{aligned}
\label{eq:update_theta_2}
\end{equation}

Since the critic works on the actor, the optimization procedure will also influence the planner's decisions. Therefore, the improvement step attempts to optimize the actor and the planner parameters $\phi, \psi$. Following \cite{haarnoja2018soft}, we can use the following objective to minimize the KL divergence between the policy and a Boltzmann distribution induced by the Q-function,

\vspace{-1em}
\begin{equation}
    \begin{aligned}
    J_{\kappa, \pi} (\psi, \phi) =& \mathbb{E}_{\mathcal{D}} \left[\alpha \log (\pi_\phi({\bf a}_t| {\bf p}_t))-Q_\theta({\bf s}_t,{\bf a}_t)\right]\\
    =&\mathbb{E}_{\mathcal{D}} \left[\alpha \log (\pi_\phi({\bf a}_t| f_\psi({\bf \epsilon}_t,{\bf s}_t)))-Q_\theta({\bf s}_t,{\bf a}_t)\right].
    \end{aligned}
\end{equation}

The last equation holds because ${\bf p}_t$ can be replaced by $f_\psi({\bf \epsilon}_t,{\bf s}_t)$ as we discussed before. 
It should be mentioned that the hyperparameter $\alpha$ can be automatically adjusted by using one proposed method from \cite{haarnoja2018soft}. Then we can apply the stochastic gradient method to optimize parameters as follows,

\begin{equation}
    \begin{aligned}
    \psi = \psi -\eta_\psi \frac{\alpha\triangledown_{{\bf p}_t}\pi_\phi({\bf a}_t| {\bf p}_t)\cdot \triangledown_{\psi} f_\psi({\bf \epsilon}_t,{\bf s}_t)}{\pi_\phi({\bf a}_t| {\bf p}_t)},
    \end{aligned}
\label{eq:update_psi_phi_1}
\end{equation}

\begin{equation}
    \begin{aligned}
    \phi = \phi-\eta_\phi\frac{\alpha\triangledown_{{\bf a}_t}\pi_\phi({\bf a}_t| {\bf p}_t)}{\pi_\phi({\bf a}_t| {\bf p}_t)}.
    \end{aligned}
\label{eq:update_psi_phi_2}
\end{equation}

\section{Meta Policy with Skip Connections}
\label{sec:su-net}

Like in a MERL model, the stochastic meta policy and maximum entropy in our framework improve the exploration for more diverse generation possibilities, which helps to prevent agents from producing a single type of plausible output during training (known as mode-collapse). 

One specific characteristic in our framework is that we also add skip-connections from each down-sampling layer of the planner to the corresponding up-sampling layer of the actor, as shown in Fig.~\ref{ovspac}. 
In this way, a natural-looking image is more likely to be reconstructed since the details of state ${\bf s}_t$ can be passed to the actor by skip-connections. Besides, since both ${\bf p}_t$ and ${\bf s}_t$ can be used by the actor to generate the executable action ${\bf a}_t$, over-exploration of the action space can be avoided in our \sw~framework, where the variance is limited by the passed detail information. 

Furthermore, the skip-connections also facilitate back-propagation of the gradients of the auxiliary learning part to the actor. 
It is also a key point to accelerate and stabilize training and avoid over-exploration since it helps the actor to focus on the refined details to bypass the coarse information from input to target.

 \bibliographystyle{elsarticle-num} 
 \bibliography{main}






\end{document}